\documentclass[10pt,journal,compsoc]{IEEEtran}

\ifCLASSOPTIONcompsoc
  \usepackage[nocompress]{cite}
\else
  \usepackage{cite}
\fi

\newcommand{\vs}{\textit{vs}. }
\newcommand{\trans}{\emph{transformer}}
\usepackage[flushleft]{threeparttable}
\ifCLASSINFOpdf
\usepackage[pdftex]{graphicx}
\DeclareGraphicsExtensions{.pdf,.jpeg,.png}
\DeclareGraphicsExtensions{.eps}
\usepackage{epstopdf}
\else
\fi

\ifCLASSOPTIONcompsoc
\usepackage[caption=false,font=footnotesize,labelfont=sf,textfont=sf]{subfig}
\else
\usepackage[caption=false,font=footnotesize]{subfig}
\fi
\usepackage{booktabs}
\usepackage{amssymb}

\usepackage{times}
\usepackage{epsfig}
\usepackage{graphicx}
\usepackage{amsmath}
\usepackage{cite}
\usepackage{enumerate}
\usepackage{cases}
\usepackage{multirow}
\usepackage{verbatim}
\usepackage{amssymb}
\usepackage{CJK}
\usepackage{algorithm}
\usepackage{algorithmicx}
\usepackage{algpseudocode}

\usepackage{color}
\usepackage{bm}
\usepackage{booktabs}
\usepackage{colortbl}
\usepackage{float}
\usepackage{ragged2e}  % use justify
\usepackage{caption}
\usepackage{csquotes}
\usepackage[breaklinks=true,bookmarks=false]{hyperref}
\usepackage{soul}

\newcommand{\tabincell}[2]{\begin{tabular}{@{}#1@{}}#2\end{tabular}} 

\newcommand{\ie}{\textit{i}.\textit{e}.}
\newcommand{\eg}{\textit{e}.\textit{g}.}

\begin{document}

\title{Contrastive Video Question Answering via Video Graph Transformer}

\author{Junbin Xiao, Pan Zhou, Angela Yao, Yicong Li, Richang Hong, Shuicheng Yan and Tat-Seng Chua \\

\IEEEcompsocitemizethanks{
\IEEEcompsocthanksitem Junbin Xiao, Angela Yao, Yicong Li and Tat-Seng Chua are with National University of Singapore. Emails: \{junbin,ayao,chuats\}@comp.nus.edu.sg, liyicong@u.nus.edu. \\
Pan Zhou and Shuicheng Yan are with Sea AI Lab, Singapore. Emails: \{zhoupan, yansc\}@sea.com \\
Richang Hong is with Hefei University of Technology, China. Email: hongrc.hfut@gmail.com \\
Corresponding to: \{junbin, chuats\}@comp.nus.edu.sg. 
}
}

% \markboth{IEEE TRANSACTIONS ON PATTERN ANALYSIS AND MACHINE INTELLIGENCES}%
% \markboth{}%
 % {Shell \MakeLowercase{\textit{et al.}}: Bare Demo of IEEEtran.cls for Computer Society Journals}

\IEEEtitleabstractindextext{%
\justify  % crucial to justify the abstract
% !TEX root = ../main.tex
%--------------------------------------------------------

\begin{abstract}
We propose to perform video question answering (VideoQA) in a \textbf{Co}ntrastive manner via a \textbf{V}ideo \textbf{G}raph \textbf{T}ransformer model (CoVGT). CoVGT's uniqueness and superiority are three-fold: 1) It proposes a dynamic graph transformer module which encodes video by explicitly capturing the visual objects, their relations and dynamics, for complex spatio-temporal reasoning. 2) It designs separate video and text transformers for contrastive learning between the video and text to perform QA, instead of multi-modal transformer for answer classification. Fine-grained video-text communication is done by additional cross-modal interaction modules. 3) It is optimized by the joint fully- and self-supervised contrastive objectives between the correct and incorrect answers, as well as the relevant and irrelevant questions respectively. With superior video encoding and QA solution, we show that CoVGT can achieve much better performances than previous arts on video reasoning tasks. Its performances even surpass those models that are pretrained with millions of external data. We further show that CoVGT can also benefit from cross-modal pretraining, yet with orders of magnitude smaller data. The results demonstrate the effectiveness and superiority of CoVGT, and additionally reveal its potential for more data-efficient pretraining. We hope our success can advance VideoQA beyond coarse recognition/description towards fine-grained relation reasoning of video contents.
Our code is available at \url{https://github.com/doc-doc/CoVGT}.
\end{abstract}

}

\maketitle

\IEEEdisplaynontitleabstractindextext

\IEEEpeerreviewmaketitle

% !TEX root = ../main.tex

%----------------------------------------------
\IEEEraisesectionheading{\section{Introduction}
\label{sec:Introduction}}
%----------------------------------VideoQA is important and Transformer is a promising solution-----------------------------------------------
\IEEEPARstart{S}{ince} 1960s, the very beginning of Artificial Intelligence (AI), continuous efforts and progress have been made towards developing AI systems that can demonstrate their understanding of the dynamic visual world by responding to the natural language queries in the context of videos which directly reflect our physical surroundings. In particular, from 2019~\cite{devlin2018bert,bommasani2021opportunities} we have been witnessing a drastic advancement in such multi-disciplinary AI, where computer vision, natural language processing, and knowledge reasoning are coordinated for accurate decision-making. The advancement stems, in part, from the success of \emph{cross-modal learning} on large-scale vision-text data \cite{radford2021learning,lu2019vilbert,su2020vl,sun2019videobert,li2021align}, and in part from a unified deep neural network for modeling of vision and natural languages, \ie, \trans~\cite{vaswani2017attention}. 
As a typical multi-disciplinary AI task, Video Question Answering (VideoQA) has benefited a lot from these developments which helps to propel the field steadily forward over the use of purely conventional techniques \cite{zhong2022video,jang2017tgif,fan2019heterogeneous,gao2018motion,jiang2020reasoning,huang2020location,xiao2021video}.

\begin{figure}[t]
  \begin{center}
    \includegraphics[width=.47\textwidth]{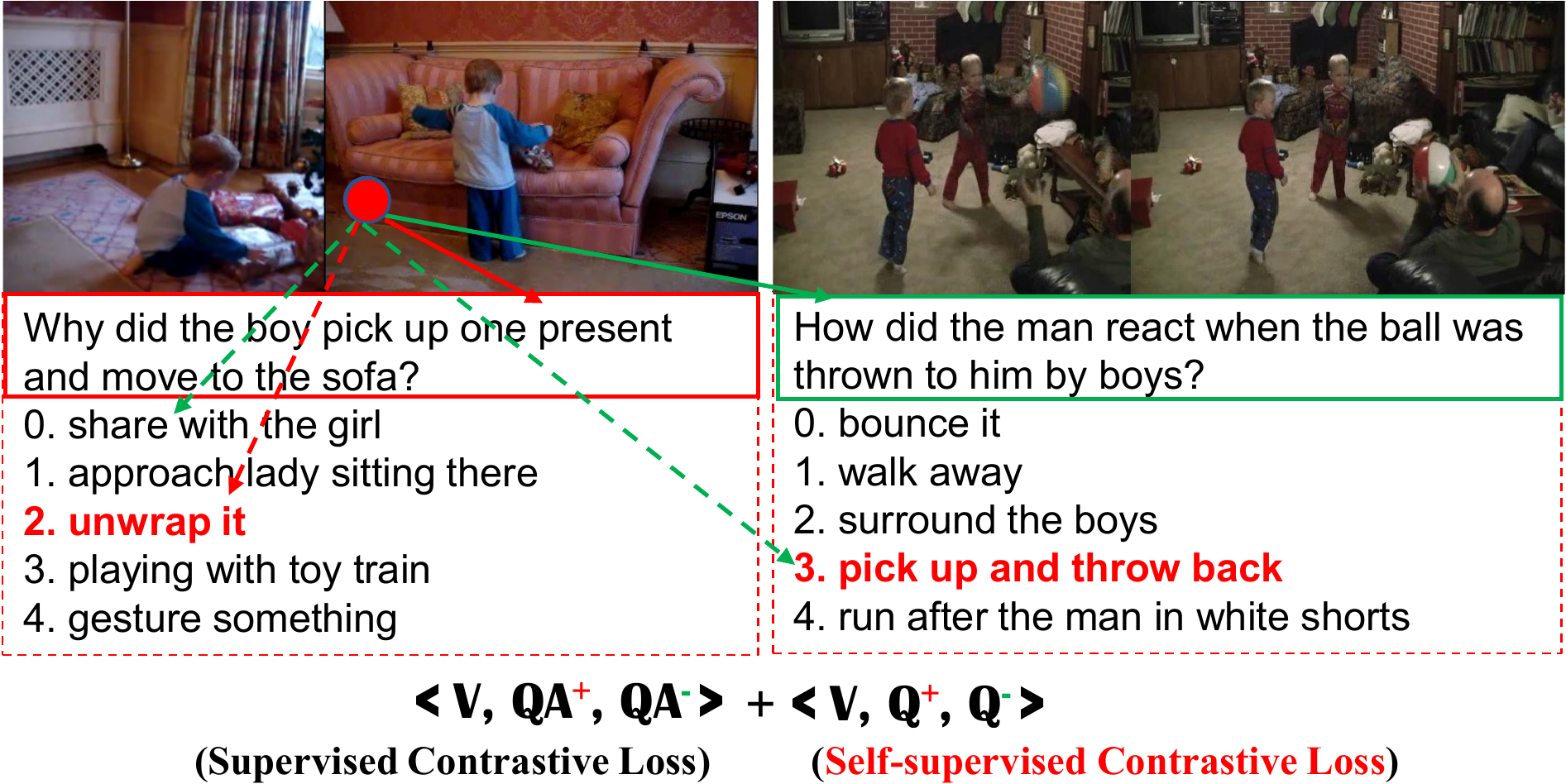}
  \end{center}
  \vspace{-0.3cm}
  \caption{Illustration of our contrastive learning strategy. For fully-supervised contrastive learning (dash arrows), we pull closer the video with its positive QA pair (red) in embedding space, while simultaneously pushing apart the video with the negative QA pairs (green). At the same time, we treat each video's question as its relevant description and collect irrelevant questions from other samples for self-supervised contrastive learning (solid arrows).}
  \label{fig:intro}
  \vspace{-0.4cm}
\end{figure}

Despite the excitement, we find that the advances made by such cross-modal learned \trans-style models mostly lie in answering questions that demand the coarse recognition or description of video contents \cite{seo2021look,xu2017video,xu2021videoclip,yang2022learning,yu2018joint,zhu2020actbert}. The problem of answering questions that challenge real-world visual relation reasoning \cite{shang2019annotating}, especially the causal and temporal relations that feature video dynamics at the action- and event-level \cite{xiao2021next,wu2021star,li2022representation}, is largely under-explored. Cross-modal pretraining seems promising \cite{lei2021less,yu2021learning,zellers2021merlot}. Yet, it requires the handling of prohibitively large-scale \emph{video}-text data \cite{fu2021violet,zellers2021merlot}, or otherwise the performances are still inferior to the state-of-the-art (SOTA) conventional techniques in handling temporal dynamics \cite{lei2021less,yu2021learning,buch2022revisiting}. In this work, we attribute the failure to the following causes:
% \begin{enumerate}

\textbf{Video encoders are overly simplistic.}
Current video encoders are either 2D or 3D neural networks operated over sparse frames~\cite{he2016deep,ren2015faster,dosovitskiy2020image} or short video segments~\cite{bertasius2021space,liu2021video,xie2018rethinking}. Such networks encode the video scene holistically, but fail to model the  spatio-temporal interactions between visual objects and reason about their compositionality \cite{xiao2021video}. Resulting models are therefore weak in reasoning, or demand learning on large-scale video data to compensate for such weak form of inputs.
    
\textbf{Formulation of VideoQA problem is sub-optimal}. Existing work solves VideoQA by classification, either cross-modal matching for multi-choice QA or multi-class classification for open-ended QA. The classification setting essentially learns a global representation (or classification layer) to predict the answer. Such a global representation is weak in disambiguating the correct versus incorrect answers. Because in multi-choice QA, the inputs of the video and question parts of a sample are the same and large, which may overwhelm the short candidate answers and dominate the overall representation. In open-ended QA, answers are treated as class indexes and their word semantics (which are helpful for QA) are not modelled. Neglecting answer semantics exacerbates the need for data and leads to sub-optimal performance as well.

\textbf{The cross-modal correspondence between video and language is insufficiently mined.} Existing models are learned on 
training samples with pure supervision oriented for question answering. Such models tend to learn spurious correlation between the inputs and the target labels. They fail to effectively capture the cross-modal correspondence (between video and language) which is essential for model to generalize  \cite{agarwal2020towards,niu2021counterfactual,li2022invariant,li2022equivariant}. Consequently, the models are prone to overfitting. This problem manifests especially in 
inference-type QA \cite{xiao2021next}, because the questions may involve multiple visual elements for multi-hop reasoning in space-time (refer to the examples in Fig.~\ref{fig:intro}).

%-----------------------------------our solution ----------------------------------------
In light of this, we propose a \textbf{Co}ntrastively learned \textbf{V}ideo \textbf{G}raph \textbf{T}ransformer model (CoVGT) that tackles the aforementioned problems and advances previous \trans-style VideoQA models in several ways. First, for the video encoder, we design a dynamic graph transformer module that explicitly captures the objects and relations as well as their dynamics to improve visual reasoning in dynamic scenarios. Second, for the problem formulation, we leverage dual-transformer architecture which maintains \emph{separate} vision and text transformers to encode video and text respectively for contrastive learning, instead of using cross-modal transformers to fuse the vision and text information for answer classification. Vision-text communication is done by additional cross-modal interaction modules. Importantly, to more effectively learn the cross-modal correspondence, we propose a joint fully- and self-supervised contrastive objective for model optimization (as illustrated in Fig.~\ref{fig:intro}). The fully-supervised contrastive objective utilizes the ground-truth answer information and enables the model to be directly optimized to distinguish the correct and incorrect answers. At the same time, the self-supervised contrastive objective captures the insight that the paired questions of a given video should be relevant to the corresponding video contents while the unpaired ones should be irrelevant. It helps to suppress common question words and
enhance the contribution of the visually-related linguistic concepts. This in turn should reduce spurious correlations between inputs and target labels during training and improve test performance.

We experiment on different VideoQA datasets that challenge the various aspects of cross-modal video understanding.
CoVGT achieves SOTA results on VideoQA benchmarks 
that challenge the reasoning of complex spatio-temporal dynamics as well as causal and commonsense knowledge
(\eg, TGIF-QA \cite{jang2017tgif}, TGIF-QA-R \cite{peng2021progressive}, NExT-QA \cite{xiao2021next}, STAR \cite{wu2021star}, and Causal-VidQA \cite{li2022representation}).
CoVGT also performs competitively on descriptive VideoQA datasets (\eg, TGIF-FrameQA\cite{jang2017tgif} and MSRVTT-QA \cite{xu2017video}) .  %achieve competitive results as well. 
Notably, CoVGT's strong performance
does \emph{not} require external data to pretrain, though pretraining with a small amount of data additionally increases accuracy. % we can observe further accuracy increase. 
The results clearly demonstrate CoVGT's effectiveness and superiority. 

This paper extends our preliminary work VGT \cite{xiao2022video} in four major aspects: 1) We propose to learn the VGT model in a joint supervised and self-supervised contrastive manner rather than merely supervised contrastive learning. The enhanced objective brings steady performance improvements by making full use of cross-modal information that are available in the VQA datasets. 2) We explore superior model implementations to realize the architecture of video graph transformer (VGT). For instance, we find that by substituting BERT \cite{devlin2018bert} with RoBERTa \cite{liu2019roberta} for language encoding can improve the performances in most cases though with small sacrifice of efficiency. Moreover, we find that the multi-choice QA performances can be improved by adding some randomness to the negative answers. 3) We substantially extend our experiments to more datasets that target at cross-modal reasoning the various aspects of video contents, \eg, STAR \cite{wu2021star} for real-world situation reasoning and Causal-VidQA \cite{li2022representation} for both evidence and commonsense reasoning. 4) We comprehensively analyze our model's strength and the contribution of each component. Furthermore, we share some heuristic observations about the performances of pretraining of visual graph transformer on cross-modal video reasoning tasks. For example, we find that existing cross-modal pretraining can hardly improve action- and event-level temporal relation reasoning in videos, which calls for more future efforts towards this direction.

% To summarize our contributions:
Our contributions are summarized as follows:
\begin{enumerate}
    \item We propose a contrastively learned video graph transformer model (CoVGT) to advance VideoQA from coarse recognition (or description) to fine-grained visual relation reasoning in dynamic scenarios. The model achieves SOTA results on a wide range of video reasoning tasks.
     \item We propose a dynamic graph transformer module which jointly models the objects, their relations and dynamics, for visual reasoning. The module is task-agnostic and can be applied to other video-language tasks.
    \item We propose to solve VideoQA in a joint fully-supervised and self-supervised contrastive manner by mining the distinction between correct and incorrect answers, as well as the relevant and irrelevant questions of a given video. Such a strategy is the first of its kind in VideoQA and shows steady advantages across different benchmarks. 
    \item To our best knowledge, we are the 1st to perform pretraining on visual graph transformer for video-language understanding (VLU). We hope that our success could promote VLU towards a more fine-grained and data-efficient direction.
\end{enumerate}

% !TEX root = ../main.tex
%--------------------------------------------------------
\section{Related Work}
\label{sec:rw}
\subsection{Conventional Techniques for VideoQA}
% In deep-learning based VideoQA (
Prior to Transformer's success for vision-language tasks, various techniques such as cross-modal attention \cite{jang2017tgif,li2019beyond,jiang2020divide}, motion-appearance memory \cite{gao2018motion,fan2019heterogeneous,liu2021hair} and graphs \cite{jiang2020reasoning,park2021bridge} have been applied to model informative contents from video for answering questions. Nonetheless. most of these techniques are built upon a sequence of frame-level or clip-level video representations which are insufficient for fine-grained object relation reasoning. More recently, graphs that exploit object-level representations \cite{dang2021hierarchical,huang2020location,liu2021hair,peng2021progressive,seo2021attend,xiao2020visual,xiao2021video,xiao2022rethinking} have demonstrated superior performance, especially on benchmarks that emphasize relation reasoning between objects and actions \cite{jang2017tgif,xiao2021next}. 
However, these graph-based methods build either monolithic graphs \cite{huang2020location} that do not disambiguate between relations in 1) space and time and 2) local and global, or static graphs at frame-level without explicitly capturing the temporal dynamics \cite{xiao2021video,peng2021progressive,liu2021hair}. The monolithic graph is cumbersome to adapt to long videos where multiple objects interact in space-time, while the static graph may lead to incorrect relations (\eg, \texttt{hug} \vs \texttt{fight}) or fail to capture dynamic relations (\eg, \texttt{take away}). 

In contrast to monolithic and static graphs, we maintain a local to global hierarchical graph architecture which reflects the intrinsic structure of video contents, and meanwhile we design temporal graph transformer to explicitly capture the graph dynamics over time. Moreover, we integrate strong language models and explore both fully-supervised and self-supervised contrastive learning strategy rather than a simple classification, to learn the structured video representations. 

\subsection{Transformers for VideoQA}
Transformer is a nascent technique for VideoQA. 
% though it has made big impact for TextQA \cite{devlin2018bert,liu2019roberta} and recently shows its success for ImageQA \cite{devlin2018bert,su2020vl,tan2019lxmert}. 
Several pioneering works \cite{seo2021look,yang2021just,zhu2020actbert} learn QA-favoured \trans-style models from HowTo100M \cite{miech2019howto100m} via designing proxy tasks (\eg, masked language modeling \cite{devlin2018bert} and cross-model matching \cite{radford2021learning,zhu2020actbert}), or curating dedicated supervisions (\eg, future utterance \cite{seo2021look} and QA pairs \cite{yang2021just}). Despite their stronger performance over conventional models \cite{gao2018motion,fan2019heterogeneous,liu2021hair,jiang2020reasoning,park2021bridge}, these works focus on answering questions that require only the recognition \cite{xu2017video} or shallow description \cite{yu2018joint} of the video scenes; their performances on visual relation reasoning \cite{jang2017tgif,xiao2021next,wu2021star} remains unknown. In addition, due to the heavy noise~\cite{amrani2021noise,miech2020end} and limited data scope (instructional videos) of
HowTo100M, models trained on it are weak in handling open-domain texts \cite{bain2021frozen,zellers2021merlot}. 

%The heavy noise level  and of HowTo100M, these models suffer from performance lose and unable to answer open-domain questions \cite{bain2021frozen,zellers2021merlot}. 

Recent efforts tend to leverage open-domain vision-text data for representation learning. ClipBERT \cite{lei2021less} takes advantage of human-annotated clean (to be distinguished from user-generated which is noisy) image-caption data \cite{lin2014microsoft,krishna2017visual} for pretraining. The pretraining may benefit spatial object recognition and visual-content related language understanding. Yet, gains in temporal reasoning~\cite{jang2017tgif} are limited, because temporal relations are hard to learn from static images. Furthermore, the pretraining relies on clean annotations which are expensive to obtain and hard to scale up. Yet, ClipBERT %'s end-to-end learning 
has inspired works \cite{fu2021violet,zellers2021merlot} to take advantage of user-generated vision-description data (vastly available on the Web) \cite{sharma2018conceptual,bain2021frozen,zellers2021merlot} for end-to-end learning. While promising, it is computationally expensive to end-to-end learn from raw videos on such large-scale datasets. Corresponding models, if not pretrained, are prone to over-fit the target datasets, given the complex reasoning tasks defined over videos \cite{xiao2021next,wu2021star} and the scarcity of annotated training data. 

Overall, the \emph{poor dynamic reasoning} and \emph{data-hungry} problems in existing \trans-style video-language models largely motivate this work. To alleviate these problems, we explicitly model the objects and relations for dynamic visual reasoning and incorporate structure priors (or relational inductive bias \cite{battaglia2018relational}) into transformer architectures to reduce the demand for data.

\subsection{Transformer Over Visual Graph}
While graph and transformer techniques have gained increasing attention \cite{kreuzer2021rethinking,wang2021tcl,ying2021transformers,yun2019graph} in modelling natural graph data (\eg~social connections), their combination in the video domain is still sparse.
Two recent works~\cite{cherian2022,geng2021dynamic} have explored graph transformers for video-language tasks. \cite{geng2021dynamic} focuses on video dialogues and simply applying a global transformer over pooled graph representations built from static frames to represent a video. \cite{cherian2022} proposes a tailored-made similarity-kernel in the self-attention blocks to capture the proximity of nodes in a pseudo 3D space. Both works do not explicitly take advantage of the objects and relations in adjacent frames to regulate the scene graph obtained at a static frame. Moreover, they neglect the local and global nature of video contents. In our work, we handle these problems by applying transformers global-locally at different-levels of graph granularity (\eg, nodes, edges and graphs) without disturbing the original structure of transformer, which brings better performance for video question answering.

\subsection{Contrastive Video-Language Understanding}
Contrastive learning aims to automatically learn generalizable data representations by contrasting the similar data against the dissimilar ones in the embedding space. The idea has recently shown great success for cross-modal pretraining \cite{miech2020end,kim2021self,radford2021learning,xu2021videoclip,yang2021just}. A handful of recent works have specially studied contrastive learning for VQA. They either learn to contrast the original and perturbed samples (obtained by masking~\cite{liang2020learning}
%(obtained by masking strategy \cite{liang2020learning} or paraphrase method \cite{kant2021contrast}) 
or paraphrasing~\cite{kant2021contrast}) for robust \emph{image} question answering, 
or aim to \emph{pretraining} a good model initialization to improve QA performance\cite{kim2021self,Yang_2021_ICCV}. 
Our use of contrastive learning differs from previous arts in two major aspects.  First, we harmoniously 
integrate the supervised and self-supervised contrastive objectives into one learning framework, in which the supervised contrastive objective is directly oriented for question answering and the self-supervised objective aims to enhance the cross-modal correspondence learning. Second, our method does not require to additionally curate contrastive data inputs; instead, we sample hard negatives from existing samples which is more convenient and shows effectiveness as well.

% !TEX root = ../main.tex
%--------------------------------------------------------
\section{Methodology}
\label{sec:method}
\subsection{Overview of Contrastive Solution}
Recall that our goal is to contrastively learn a video graph transformer model for VideoQA. To begin with, 
we formally define the VideoQA task as follows: given a video \textit{v} and a question \textit{q}, VideoQA aims to combine the two stream information \textit{v} and \textit{q} to predict the answer \textit{$a$}. Depending on the task setting, \textit{$a$} can be given as multiple choices along with each question for multi-choice QA, or it is defined in a global answer set for open-ended QA. 
\begin{figure}[t]
  \begin{center}
    \includegraphics[width=.47\textwidth]{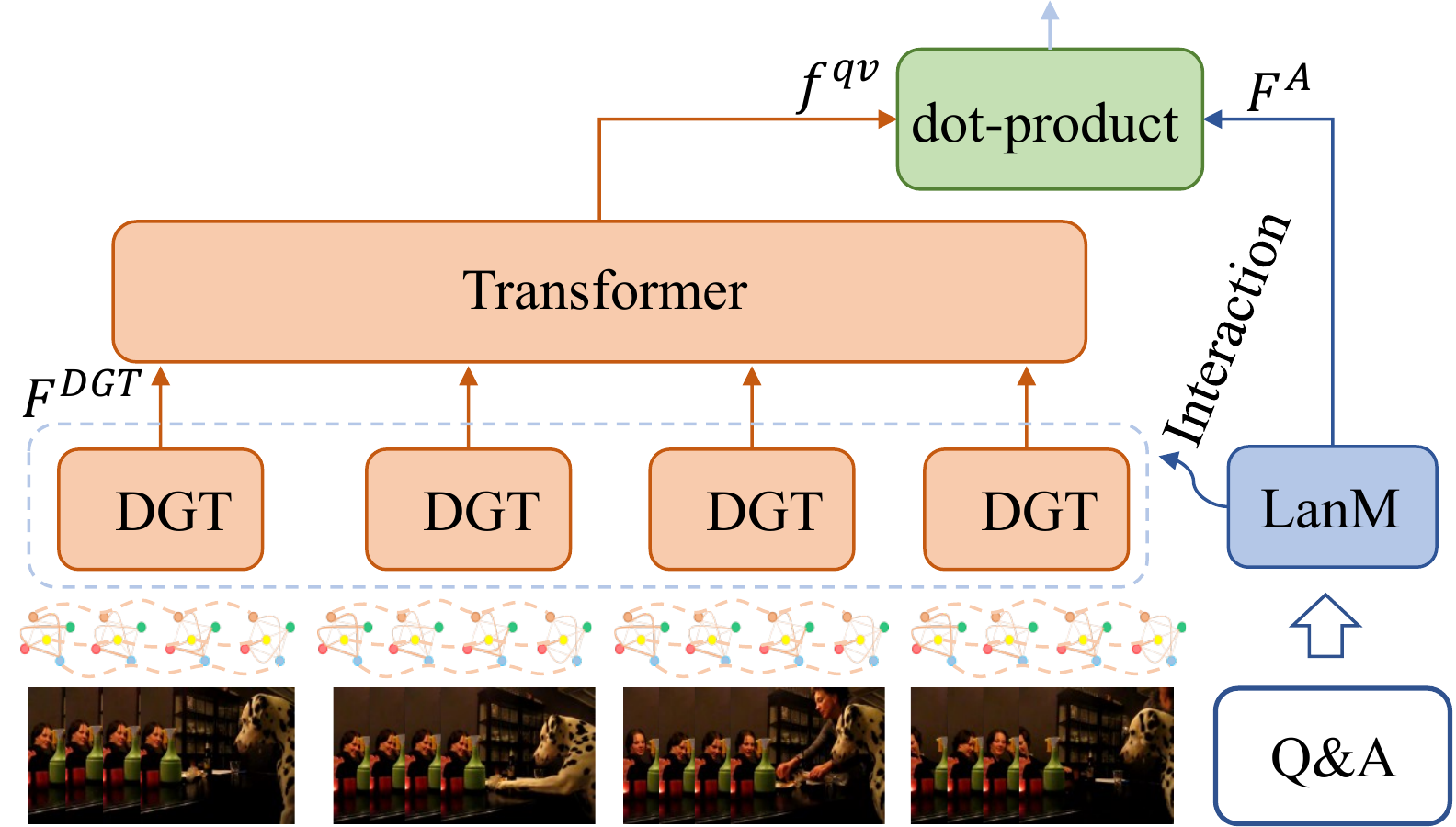}
  \end{center}
  \vspace{-0.3cm}
  \caption{Framework of video graph transformer (VGT). It projects the videos and texts into embedding space via a video encoder (orange) and a language model (blue) respectively, and computes their dot-product similarity for decision making. The video encoder maintains a local to global hierarchical architecture which includes a dynamic graph transformer (DGT) module and a global transformer. Additionally, a cross-modal interaction module is integrated to pinpoint the informative visual contents from videos.}
  \vspace{-0.4cm}
  \label{fig:framework}
\end{figure}

In this work, we handle both types of VideoQA by jointly optimizing the supervised and self-supervised contrastive objectives, \ie, for multi-choice QA:
\begin{equation}\label{eq:obj-mc}
    \mathcal{L} = \underbrace{\mathcal{L}_{vqa}(v, qa^+, qa^-)}_{\text{supervised}} + \lambda \underbrace{\mathcal{L}_{vq}(v, q^+, q^-)}_{\text{self-supervised}},
\end{equation}
and for open-ended QA:
\begin{equation}\label{eq:obj-oe}
    \mathcal{L} = \underbrace{\mathcal{L}_{vqa}(vq, a^+, a^-)}_{\text{supervised}} + \lambda \underbrace{\mathcal{L}_{vq}(v, q^+, q^-)}_{\text{self-supervised}},
\end{equation}

\noindent where $\mathcal{L}_{vqa}$ is the supervised contrastive objective oriented for question answering, and $\mathcal{L}_{vq}$ is the self-supervised contrastive objective designed to enhance the cross-modal correspondence learning between the video and its related questions. Note that for pretraining with weakly-paired video-text data (\eg, on WebVid \cite{bain2021frozen}), only the self-supervised contrastive loss function is reserved. In both Eqn.~\eqref{eq:obj-mc} and Eqn.~\eqref{eq:obj-oe}, $vq$ and $qa$ denote the combined (by concatenation) inputs of the video and question, question and answer respectively. $\mathcal{L}$ is the loss value, $\lambda$ is a trade-off parameter and $\mathcal{L}_*(x, x^+, x^-)$ is given by the InfoNCE loss \cite{oord2018representation}:
\begin{equation}
    -\mathbb{E}_i [\log(\frac{e^{s_{\text{VGT}}(x_i,x_i^+)}}{e^{s_{\text{VGT}}(x_i,x_i^+)}+\sum\limits_{(x_i, x_j^-)\in \mathcal{N}_i} e^{s_{\text{VGT}}(x_i, x_j^-)}})],
\end{equation}
where $x$, $x^+$, $x^-$ denote the placeholders for the inputs of the anchor, positive and negative samples respectively.  $s_{\text{VGT}}$ denotes the dot-product value of the two inputs in the embedding space. The embeddings are computed by our video graph transformer (VGT) model illustrated in Fig.~\ref{fig:framework}. Specifically: 
\begin{equation}
    s_{\text{VGT}}(x, x^+) = \mathcal{F}_{cm}(\mathcal{F}_{\text{vid}}(x), \mathcal{F}_\text{lang}(x^+))^\intercal \cdot \mathcal{F}_\text{lang}(x^+), 
\end{equation}
where $\mathcal{F}_{cm}$, $\mathcal{F}_{\text{vid}}$ and $\mathcal{F}_{\text{lang}}$ denote the cross-modal interaction module, the video encoder and language encoder respectively. The parameters to be optimized are contained in these three modules.

Our solution differs from the vast majority of previous works that formulate and solve VQA as a classification problem~\cite{jang2017tgif,lei2021less,xiao2021video}. Instead, we formulate VQA as a cross-modal contrastive learning problem and explicitly optimize the distinction between the correct and incorrect answers, as well as the relevant and irrelevant questions with respect to a given video. Our solution enjoys several major advantages: For multi-choice QA, it encodes video and QAs separately with different transformers (\eg, $\mathcal{F}_{\text{vid}}$ and $\mathcal{F}_{\text{lang}}$), and thus circumvents the over-fitting and hard-negative answer issues resulted from representing each $\langle\text{video}, \text{question}, \text{candidate answer}\rangle$ triplet with a single feature from cross-modal transformer for classification. As a result, it can achieve much better performances. For open-ended QA, it encodes the semantics of the answer words which can benefit question answering. Last but not least, our formulation makes it %it brings 
convenience to adapt the model architecture to different datasets, since it dispenses with the classification layer which are usually dataset-specific. As such, it can fully enjoy the pretrained weights in the \emph{pretrain-and-finetune} experiments.

We next elaborate the details of our model by first introducing the video encoder which comprises a video graph representation stage in Sec.~\ref{sec:vgr} and a dynamic graph transformer module in Sec.~\ref{sec:dgt}, and finally a global transformer in Sec.~\ref{sec:gt}. Cross-modal interaction along with text encoder is presented in Sec.~\ref{sec:cm}. Finally, we introduce the answer decoder in Sec.~\ref{sec:ad}. 

\subsection{Video Graph Representation}
\label{sec:vgr}
\begin{figure}[t]
  \begin{center}
    \includegraphics[width=.47\textwidth]{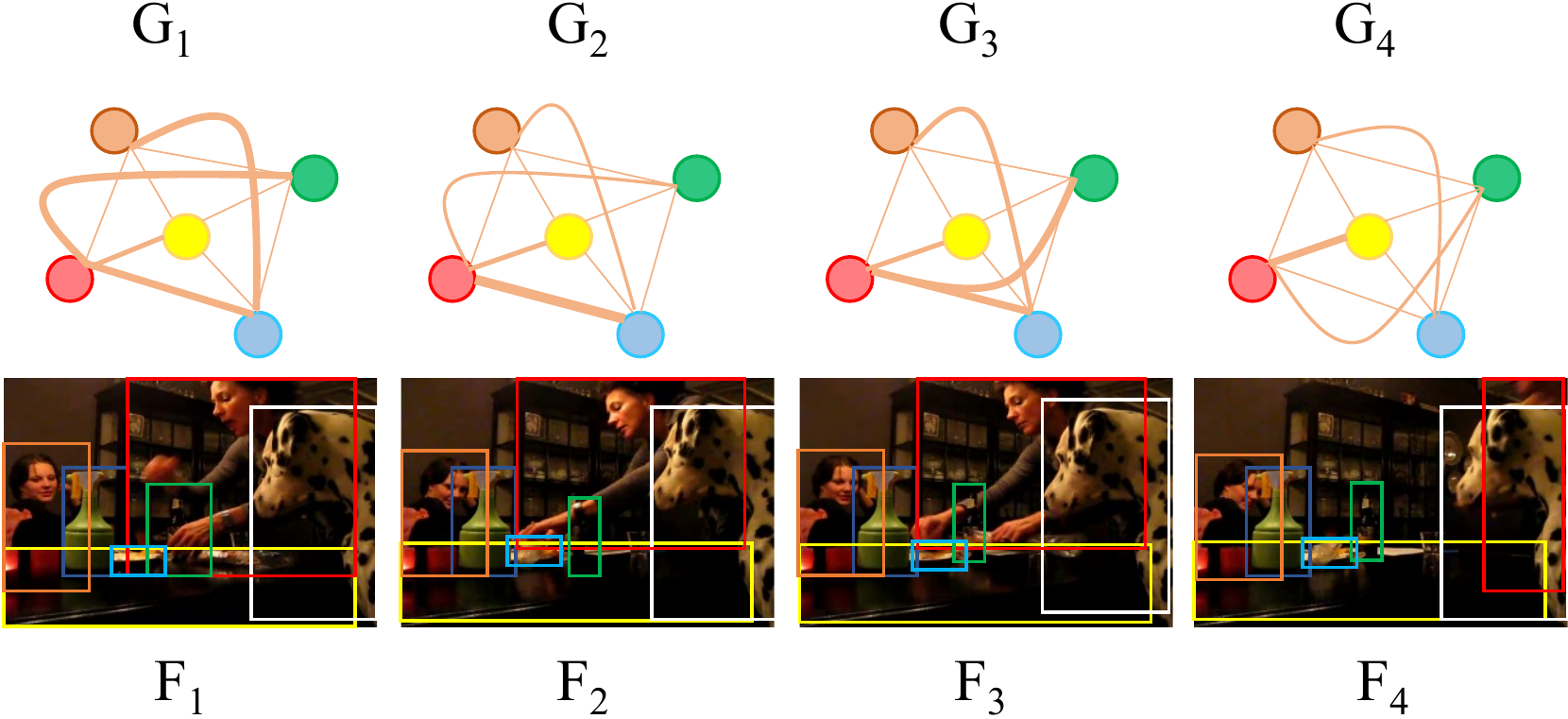}
  \end{center}
  \vspace{-0.3cm}
  \caption{Illustration of graph representation for a short video clip with 4 frames. The nodes of the same color represent the same object. Self-loops are omitted for brevity.}
%   Moreover, the feature representations of nodes and edges corresponding to the same objects vary among different frames. }
  \vspace{-0.4cm}
  \label{fig:vgr}
\end{figure}

To explicitly model the visual objects and their relationships for visual reasoning, it is necessary to represent a video as visual graphs whose nodes and edges correspond to visual objects and their relationships respectively \cite{tsai2019video,xiao2021video}.
% Visual object graphs, whose nodes and edges correspond to the visual objects and their relations respectively, have demonstrated compelling performance in object relation reasoning \cite{tsai2019video,xiao2021video}. 
In this section, we introduce in detail how to convert a video clip into visual object graphs. Given a raw video, we sparsely sample $l_v$ frames in a way analogous to \cite{xiao2021video}. The $l_v$ frames are evenly distributed into $k$ clips of length $l_c=\frac{l_v}{k}$.
For each sampled frame (see Fig.~\ref{fig:vgr}), we extract $n$ RoI-aligned features as object appearance representations $F_r\!=\!\{f_{r_i}\}_{i=1}^n$ along with their spatial locations $B\!=\!\{b_{r_i}\}_{i=1}^n$ with a pretrained object detector~\cite{anderson2018bottom,ren2015faster}, where $r_i$ represents the $i$-th object region in a frame. Additionally, we obtain an image-level feature $F_I\!=\!\{f_{I_t}\}_{t=1}^{l_v}$ for all the sampled frames with a pretrained image classification model~\cite{he2016deep}. $F_I$ will serve as the global contexts to augment the graph representations aggregated from the local objects.

To find the same object across different frames within a clip, we define a linking score $s$ by considering the appearance and spatial location:
\begin{equation}
    s_{i, j} =  \psi(f_{r_i}^t, f_{r_j}^{t+1}) + \gamma*\text{IoU}(b_i^t, b_j^{t+1}),
\end{equation}
where $\psi$ computes the appearance similarity between detected objects $i$ and $j$ in two adjacent frames.  In this work, we use the cosine similarity for $\psi$, while IoU denotes intersection-over-union between the bounding boxes of objects $i$ and $j$. The trade-off hyper-parameter $\gamma$ is set to 1 in our experiments. The $n$ detected objects in the first frame of each clip are designated as anchor objects. Detected objects in consecutive frames are then linked to the anchor objects by greedily maximizing $s$ frame by frame\footnote{We assume that the group of objects do not change in a short video clip.}. By aligning the objects within a clip, we ensure the consistency of the node and edge representations for the graphs constructed at different frames (we construct one graph per frame at this stage).

After the alignment, we concatenate for each object its appearance $f_r$ and location representation $f_{loc}$, and project the combined feature into the $d$-dimensional space via 
\begin{equation}
\label{equ:fo}
    f_o = \mathrm{ELU}(\phi_{W_o}([f_r;f_{loc}])),
\end{equation}
where $[;]$ denotes concatenation and $f_{loc}$ is obtained by applying a $1 \times 1$ convolution over the relative coordinates as in \cite{xiao2021video}. The function $\phi_{W_o}$ denotes linear transformation with parameters $W_o$. 

After obtaining the object representations $F_o\!=\!\{f_{o_i}\}_{i=1}^n$, their relationships in the $t$-th frame can be initialized as pairwise similarities in adjacency matrix $R_t$ as follows
\begin{equation}
\label{equ:aa}
    R_t = \sigma(\phi_{W_{ak}}(F_{o_t})\phi_{W_{av}}(F_{o_t})^\mathrm{\intercal}), \quad t\in\{1, 2, \dots, l_v\},
\end{equation}
where $\phi_{W_{ak}}$ and $\phi_{W_{av}}$ denote linear transformations with parameters $W_{ak}$ and $W_{av} \in \mathbb{R}^{d \times \frac{d}{2}}$ respectively. We use different transformations to make the relation asymmetric, which reflects the nature of real-world subject-object interaction \cite{xiao2020visual,krishna2018referring} (\eg, \texttt{hit} \vs \texttt{being hit}). As for symmetric relations, we expect that their transformed representations are quite similar. Here $\mathrm{\intercal}$ indicates matrix transpose, and $\sigma$ is the Softmax operation that normalizes each row. After initialization, the values of the $i$-th row in the adjacency matrix $R_t$ denote the relations of object $i$ with regard to all of the other objects in the $t$-th frame. 

The adjacency matrix $R$ is obtained independently for each frame; hence there are $l_c$ such adjacency matrices for each video clip.  However, as the objects are aligned in the video clip, each entry of the adjacency matrices always corresponds to the relation for the same pair of objects. For brevity, we use $G_t=(F_{o_t}, R_t)$ to denote the graph representation of the $t$-th frame where $F_{o}$ are node representations and $R$ are edge representations of the graph.

\subsection{Dynamic Graph Transformer}
\label{sec:dgt}
\begin{figure}[t]
  \begin{center}
    \includegraphics[width=.4\textwidth]{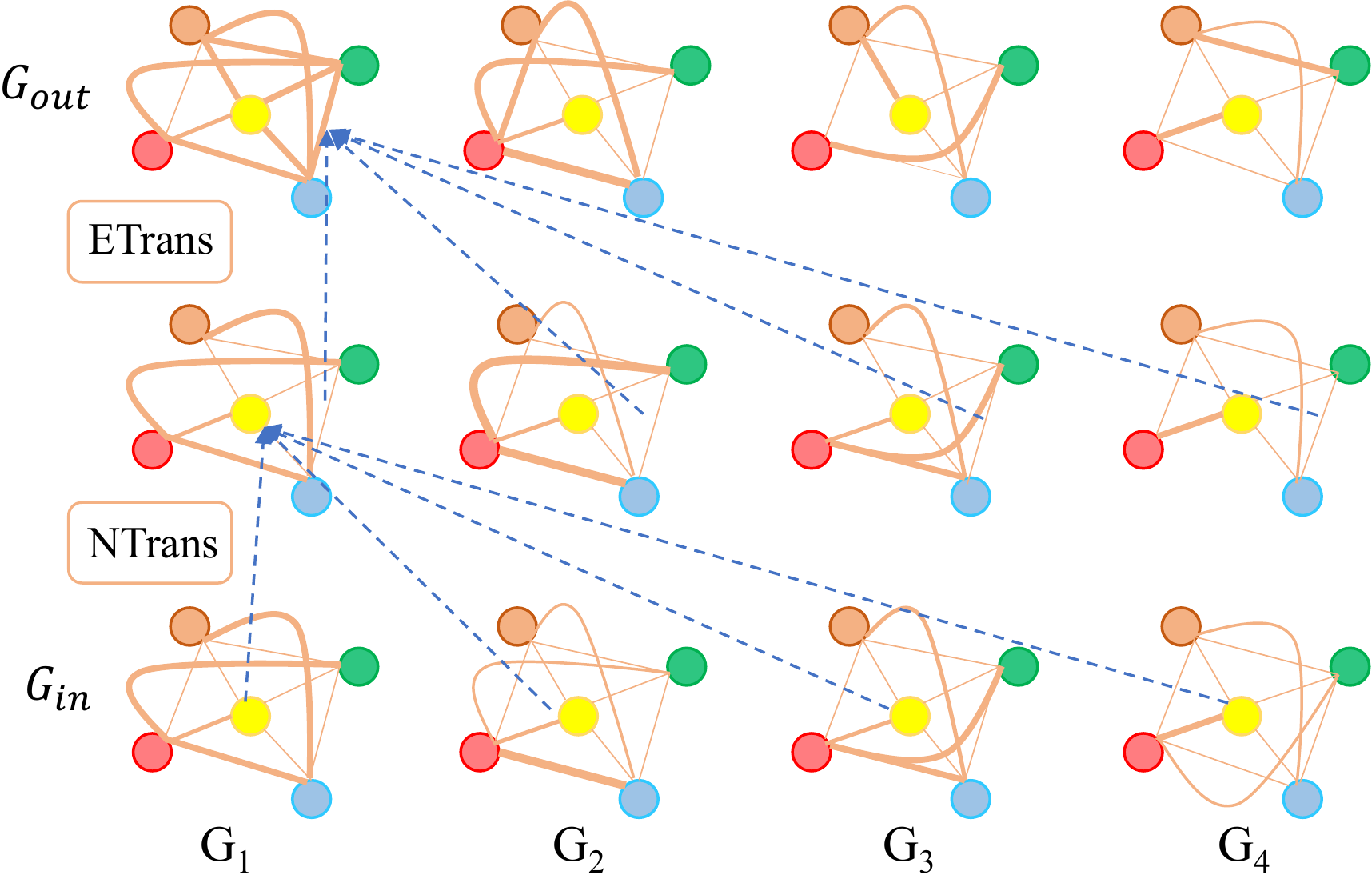}
  \end{center}
  \vspace{-0.3cm}
  \caption{Illustration of temporal graph transformer in a short video clip. It contains a node transformer (NTrans) and an edge transformer (ETrans) that are operated sequentially. }
  \vspace{-0.4cm}
  \label{fig:dgt}
\end{figure}
Previous efforts construct visual graphs at static frame level \cite{xiao2021video, peng2021progressive,liu2021hair}, thus failing to explicitly capture the object dynamics. Therefore, at the heart of our video encoder is the dynamic graph transformer (DGT) module. This module takes as inputs a set of visual graphs $\{G_t\}_{t=1}^{L_v}$ clip-wisely, and outputs a sequence of representations $F^{DGT}\in\mathbb{R}^{d\times k}$ by mining the temporal dynamics of objects and their spatial interactions. To achieve this, we sequentially operate a temporal graph transformer unit, a spatial graph convolution unit and a hierarchical aggregation unit. 

\subsubsection{Temporal Graph Transformer}
As illustrated in Fig.~\ref{fig:dgt}, the temporal graph transformer unit operates over a single video clip.  It takes as input the set of graphs $G_{in}=\{G_t\}_{t=1}^l$ from that clip and outputs a new set of graphs $G_{out}$ by mining the temporal dynamics of the objects and their relations (\ie~interactions). Specifically, there is a node transformer (NTrans) and an edge transformer (ETrans) that operate over the graph nodes and edges respectively. The two types of transformers are designed to exploit the objects and their relations in adjacent frames to enhance the object representations and calibrate the relationships captured at static frames.
For completeness, we briefly recap the self-attention mechanism in transformer \cite{vaswani2017attention}. 
Given a sequence of inputs $X_{in}=\{x_{in}^t\}_{t=1}^l$,
the transformer module mix token information by employing a multi-head self-attention (MHSA):
\begin{equation}
X_{out} = \text{MHSA}(X_{in}) = \phi_{W_c}([h_1;h_2;\dots,h_e]),
\end{equation}
where $\phi_{W_c}$ is a linear transformation with parameters $W_c$, and
\begin{equation}
    h_i = \text{SA}(\phi_{W_{i_q}}(X_{\text{in}}), \phi_{W_{i_k}}(X_{\text{in}}), \phi_{W_{i_v}}(X_{\text{in}})),
\end{equation}
where $\phi_{W_{i_q}}$, $\phi_{W_{i_k}}$ and $\phi_{W_{i_v}}$ denote the linear transformations of the query, key, and value vectors of the $i$-th self-attention (SA) head respectively. $e$ denotes the number of self-attention heads, and SA is defined as:
\begin{equation}
     \text{SA}(X_q, X_k, X_v) = \sigma\left({X_k X_q^\mathrm{\intercal}/\sqrt{d_k}}\right)X_v, 
\end{equation}
in which $d_k$ is the dimension of the key vector. Finally, a skip-connection with layer normalization (LN) is applied to the output sequence $X=LN(X_{out}+X_{in})$. The final $X$ may undergo further MHSAs depending on the number of transformer layers.

In our temporal graph transformer, we first apply $H$ number of MHSA blocks to enhance the node (or object) representations by aggregating information from other nodes of the same object from all $l$ adjacent frames within a clip:
\begin{equation}
\label{equ:node}
    F'_{o_i} = \text{NTrans}(F_{o_i})=\text{MHSA}^{(H)}(F_{o_i}),
\end{equation}
in which $F_{o_i} \in \mathbb{R}^{l \times d}$ denotes a sequence of feature representations corresponding to object $i$ in a video clip of length $l$. Our motivation behind the node transformer is that it helps model the change of single object behaviours and thus infer the dynamic actions (\eg~\texttt{bend down}). Also, it is helpful in improving the appearance features in the cases where the object at certain frames suffer from motion blur or partial occlusion.

Based on the new node representations $F'_o=\{F'_{o_i}\}_{i=1}^n$, we can update the relation matrix $R$ via Eqn.~\eqref{equ:aa}. Then, to explicitly model the temporal dynamics of the relations, we further apply a edge transformer (of $H$-MHSA layers) on all the updated relation matrices: 
\begin{equation}
\label{equ:edge}
    \mathcal{R'} =\text{ETrans}(\mathcal{R}) =\text{MHSA}^{(H)}(\mathcal{R}),
\end{equation}
where $\mathcal{R}\!=\!\{R_t\}_{t=1}^l\in\mathbb{R}^{l \times d_n}$ ($d_n=n^2$) is the $l$ adjacency matrices that are row-wisely expanded. ffHere, our motivation is that the relations captured at the static frames may be incorrect (\eg, \texttt{hug} \vs \texttt{fight}), trivial (\eg, \texttt{tough} \vs \texttt{wipe}) or incomplete (\ie, unable to identify dynamic relations like \texttt{put down}), and the edge transformer can help to calibrate the wrong relations and recall the missing ones. Note that in our implementation, the temporal positions for both node- and edge- transformers are not used since the transformers are applied within a short video clip, and we empirically find that the temporal positions do not help the performance. For brevity, we refer to the resultant graph at the $t$-th frame as $G_{out_t}=(F'_{o_t}, R'_t)$. 

\subsubsection{Spatial Convolution and Hierarchical Aggregation}
\begin{figure}[t]
  \begin{center}
  \scalebox{0.5}{
    \includegraphics[width=.35\textwidth]{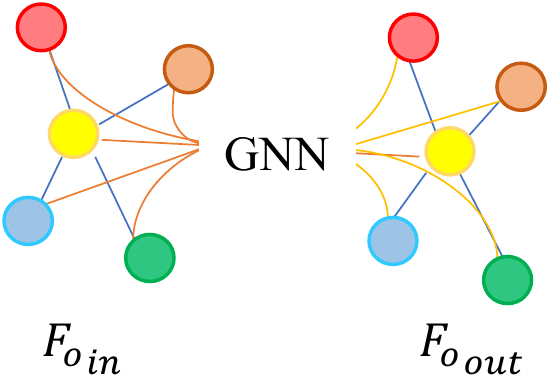}
    }
  \end{center}
  \vspace{-0.3cm}
  \caption{Illustration of spatial graph convolution. Given a set of input nodes $F_{o_{in}}$, this unit applies GNN to enhance the node representations according to their relations with regard to their neighbors. The nodes denote the visual objects.}
%   \vspace{-0.4cm}
  \label{fig:gr}
\end{figure}
\textbf{Spatial Graph Convolution}. The temporal graph transformer focuses on temporal relation capturing. To reason over the object spatial interactions, we apply a $U$-layer graph attention convolution \cite{kipf2016semi} (as illustrated in Fig.~\ref{fig:gr}) on all the $l_v$ graphs:
\begin{equation}
\label{equ:gcn}
    F^{'(u)}_o = \text{ReLU}((R'+I)F_o^{'(u-1)} W^{(u)}),
\end{equation}
in which $W^{(u)}$ denote the graph parameters at the $u$-th layer. $I$ is the identity matrix for skip connections. $F^{'(u)}_o$ are initialized with $F'_o$ which are the output node representations of the temporal graph transformer unit. The index $t$ is omitted for brevity. Afterwards, a last skip-connection: $F_{o_{out}}=F'_o+F_o^{'(U)}$ is used to obtain the final node representations. 

\begin{figure}[t]
  \begin{center}
  \scalebox{0.5}{
    \includegraphics[width=0.7\textwidth]{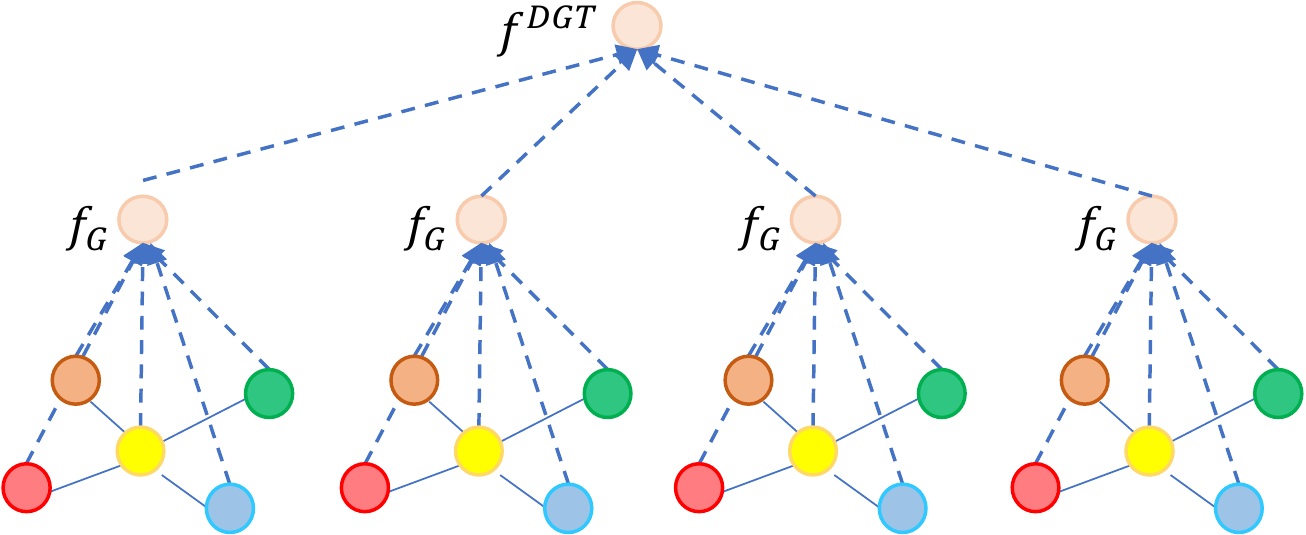}
    }
  \end{center}
  \vspace{-0.3cm}
  \caption{Illustration of hierarchical aggregation.}
  \label{fig:hpool}
  \vspace{-0.4cm}
\end{figure}

\noindent \textbf{Hierarchical Aggregation}.
The node representations so far have explicitly taken into account the objects' temporal dynamics and spatial interactions. Nonetheless, such interactions are mostly atomic. 
%(\eg, $\langle$\texttt{boy stretches out hand}, \texttt{boy grabs toy pieces}, \texttt{boy plugs pieces}$\rangle$ \vs \texttt{boy assembles toy}). 
To aggregate the atomic interactions into higher level video elements and to narrow the semantic gap between the visual and textual representations, we design a hierarchical aggregation strategy as illustrated in Fig.~\ref{fig:hpool}. First, we aggregate the graph nodes at each frame by a simple self-attention:
\begin{equation}
   f_G = \sum\nolimits_{i=1}^N \alpha_i F_{o_{out_i}}, \quad \text{where} \quad \alpha = \sigma(\phi_{W_G}(F_{o_{out}})),
\end{equation}
and $\phi_{W_G}$ is a linear transformation with parameters $W_G \in \mathbb{R}^{d \times 1}$. 
The graph representation usually captures local object interactions. However, it may lose sight of the global picture of a frame, especially since we only detect $n$ objects and cannot guarantee that the detected objects will contain all the objects of interest in that frame. To enhance graph representation, we concatenate $F_G$ with the frame-level appearance features $F_a$:
\begin{equation}
\label{equ:ff}
    f_G = \text{ELU}(\phi_{W_m}([\phi_{W_f}(f_a);f_G]))
\end{equation}
in which $\phi_{W_m}$ and $\phi_{W_f}$ are linear transformations with parameters $W_m\in\mathbb{R}^{2d \times d}$ and $W_f\in\mathbb{R}^{2048\times d}$ respectively.

We next aggregate the local interactions to obtain a sequence of clip-level feature representations via a simple mean-pooling:
\begin{equation}
\label{equ:fc}
 f^{\text{DGT}}=\text{MPool}(F_G)=\frac{1}{l_c}\sum\limits_{t=1}^lf_{G_t}
\end{equation}
\noindent The set of $k$ clips are finally represented by  $F^{\text{DGT}}\!=\!\{f_c^{\text{DGT}}\}_{c=1}^k$. 

\subsection{Global Transformer}
\label{sec:gt}
The aforementioned DGT module pays attention to derive informative visual clues from the local video contents. To capture the temporal and causal relations between these local video contents, we employ another $H$-layer transformer over the outputs of DGT (\ie~$F^{\text{DGT}}$), and add learnable sinusoidal temporal position embeddings~\cite{devlin2018bert}. Finally, the transformer's outputs are mean-pooled to obtain the global representation $f^v\in\mathbb{R}^d$ for the entire video. This process is formally represented by
\begin{equation}
\label{equ:gbtrans}
    f^v = \text{MPool}(\text{MHSA}^{(H)}(F^{\text{DGT}})).
\end{equation}
The global transformer has two major advantages. First, it retains the overall hierarchical structure which progressively drives the video elements at different granularity as in \cite{xiao2021video}. Second, it enhances the compatibility of the visual and text representations which may benefit cross-modal comparison.
\setlength{\tabcolsep}{9pt}
\begin{table*}[t!]
\small
\centering
\caption{Dataset statistics. OE-1450: open-ended QA with 1450 global answer candidates. MC-5: multi-choice QA with 5 options and only one of them is correct. Note that TGIF-QA-R \cite{peng2021progressive} shares the same statistics with TGIF-QA.}
\label{tab:dset}
\vspace{-0.5em}
\begin{threeparttable}
    \scalebox{0.8}{
    \begin{tabular}{llcccccc}
        \hline\hline
        Datasets & Main Challenges & \#Videos/\#QAs & Train & Val & Test & Video Length (s) & QA \cr
        \hline
        NExT-QA \cite{xiao2021next} & Causal \& Temporal Reasoning  & 5.4K/48K & 3.8K/34K & 0.6K/5K & 1K/9K & 44 & MC-5 \cr
        
         \multirow{3}*{TGIF-QA \cite{jang2017tgif}} 
         & Repetition Action  & 22.8K/22.7K & 20.5K/20.5K & -& 2.3K/2.3K & 3 & MC-5\cr
         & State Transition  & 29.5K/58.9K & 26.4K/52.7K & -& 3.1K/6.2K & 3 & MC-5\cr
         & Frame QA         & 39.5K/53.1K & 32.3K/39.4K & -& 7.1K/13.7K & 3 & OE-1450\cr
        %  \cline{1-8}
         STAR-QA \cite{wu2021star} & Situated Reasoning  & 5K/ 60K & 3K/46K & 1K/7K & 1K/7K & 30& MC-4\cr
        %  \cline{1-8}
          Causal-VidQA \cite{li2022representation} & Evidence \& Commonsense Reasoning  & 26.9K/ 161.4K & 18.8K/112.7K & 2.7K/16.0K & 5.4K/32.6K & 9 & MC-5\cr
        %  \cline{1-8}
         MSRVTT-QA \cite{xu2017video} & Visual Recognition  & 10K/ 244K & 6.5K/159K & 0.5K/12K & 3K/73K & 15& OE-4000\cr
        
        \hline
    \end{tabular}
    }
    \vspace{-0.3cm}
\end{threeparttable}
\end{table*}
\subsection{Cross-modal Interaction}
\label{sec:cm}
To find the informative visual contents with respect to a particular text query, a cross-model interaction between the visual and textual nodes is designed. Given a set of visual nodes represented by $X^v$, we incorporate textual information represented by $X^q=\{x_m^q\}_{m=1}^M$ (M denotes number of tokens in the text query) into the visual nodes via
\begin{equation}
x^{qv} =\mathcal{F}_{cm}(x^v, X^q) = x^v + \sum\nolimits_{m=1}^M \beta_m x^q_m,
\end{equation}
where $X^v$ and $X^q$ denote the placeholders for the visual and text representations for interaction module. $\beta = \sigma((x^v)^\intercal X^q)$. The query-aware nodes $X^{qv}$ then substitute the original visual nodes and proceed to subsequent operations. 

In principle, this cross-modal interaction module can be applied at different level of visual abstractions (object-level $F_O$ in Eqn.~\eqref{equ:fo}, frame-level $F_G$ in Eqn.~\eqref{equ:ff} or clip-level $F^{DGT}$ in Eqn.~\eqref{equ:fc}) to be in line with the multi-granularity of video elements and linguistic concepts~\cite{xiao2021video}.
% though results may be dataset dependent. 
In our implementation, we investigate several variants and find that performances vary among different datasets.
As a default, we simply plug it at the outputs of the DGT module (\ie~$X^v:=F^{\text{DGT}}$) for efficiency consideration as the number of visual representations at this stage is much smaller. Empirically, this implementation generalizes better across different datasets, because the visual nodes at this level have already absorbed the information of both static and dynamic from preceding layers. For the textual node $X^q$, we obtain them by a simple linear projection on the token outputs of a language model (\eg, $\mathcal{F}_{\text{lang}}$ can be BERT \cite{devlin2018bert} or RoBERTa \cite{liu2019roberta}):
\begin{equation}
\label{equ:bproj}
    X^q=\phi_{W_Q}(\mathcal{F}_{\text{lang}}(Q)),
\end{equation}
where $W_Q\in\mathbb{R}^{768 \times d}$, and Q denotes question words in open-end QA or words of QA pairs in multi-choice QA. Particularly, in multi-choice QA, the candidate answers of a question will be appended to the question to form multiple textual queries. In this scenario, we max-pool the obtained query-aware visual representations with respect to different QA queries to find the one that is mostly relevant to the video. Our intuition is that in most cases, the QA queries corresponding to the correct answers are mostly relevant to the video. 

\subsection{Answer Prediction}
\label{sec:ad}
To get a global representation for a particular answer candidate, we mean-pool its token representations from the language model: 
\begin{equation}
\label{equ:mpool}
    f^A = \text{MPool}(X^A),
\end{equation}
where $X^A$ denotes a candidate answer's token representations, and is obtained by feeding the text answer to the language model in a way analogous to Eqn.~\eqref{equ:bproj}. Its similarity with the query-aware video representation $f^{qv}$ (obatined via Eqn.~\eqref{equ:gbtrans}) is thus obtained through a dot-product between the two vectors. Consequently, the candidate of maximal similarity is returned as a prediction:
\begin{equation}
    a^* = \arg \max((f^{qv})^\intercal F^A),
\end{equation}
in which $F^A=\{f^A_a\}_{a=1}^{|\mathcal{A}|}\in\mathbb{R}^{d \times |\mathcal{A}|}$, and $|\mathcal{A}|$ means the number of answer candidate.
Additionally, for open-ended QA, we follow previous works \cite{xiao2021video} and enable a video-absent QA by directly computing the similarities between the question representation $f^q$ (obtained in a way similar to $f^A$.) and the answer representations $F^A$. As a result, the final answer can be a joint decision: 
\begin{equation}
\label{equ:oe}
    a^* = \arg \max((f^{qv})^\intercal F^A \odot (f^q)^\intercal F^A),
\end{equation}
in which $\odot$ is element-wise product. 

% !TEX root = ../main.tex

%--------------------------------------------------------
\section{Experiments}
\label{sec:exp}

\subsection{Datasets}
We experiment on different datasets. Seven datasets (NExT-QA \cite{xiao2021next}, TGIF-Action and Transition \cite{jang2017tgif}, TGIF-QA-R Action and Transition \cite{peng2021progressive}, STAR-QA \cite{wu2021star} and Causal-VidQA \cite{li2022representation}) challenge the complex temporal and causal relation as well as commonsense reasoning in videos (feature temporal dynamics). Two additional datasets (TGIF FrameQA \cite{jang2017tgif} and MSRVTT-QA \cite{xu2017video}) challenge the recognition of the video objects, their attributes and actions as well as activities (feature frame statics).
The related statistics of the datasets are presented in Tab.~\ref{tab:dset}. Other details are given in the Appendix \ref{app:dataset}. For all experiments, we follow standard protocol and report accuracy (percentage of correctly answered questions) for evaluation metric.

\setlength{\tabcolsep}{7.8pt}
\begin{table*}[t]
    \small
    \centering
    \caption{Accuracy (\%) comparison on NExT-QA \cite{xiao2021next}. F: Frame-level feature from ResNet or ViT or their variants. C: Clip-level feature from 3D neural networks. C$^+$: cross-modal pretrained S3D \cite{miech2020end}. R: Region-level feature from Faster R-CNN. Acc@C, T, D, denote accuracy for Causal, Temporal, and Descriptive questions respectively. *: results reproduced with the official code. The \textbf{best} and \underline{2nd best} results are highlighted in bold and underline respectively.
    }
    \vspace{-0.5em}
    \scalebox{0.8}{
    \begin{tabular}{l|cc|c|c|ccc|c|ccc|c}
    \hline\hline
    \multirow{3}*{Methods} & \multicolumn{2}{c|}{\multirow{1}*{Pretrain}}& \multirow{3}*{Video} & \multirow{3}*{Text} & \multicolumn{4}{c|}{NExT-QA Val} & \multicolumn{4}{c}{NExT-QA Test} \\ \cline{2-3} \cline{6-13}
    ~ & \multirow{2}*{Dataset} & \multirow{2}*{Size} & ~ & ~ & Acc@C & Acc@T & Acc@D & \multirow{2}*{Acc@All} & Acc@C & Acc@T & Acc@D & \multirow{2}*{Acc@All} \\ 
     ~ & ~ & ~ & ~ & ~ & (48\%) & (29\%) & (23\%) & ~ & (48\%) & (29\%) & (23\%) & ~ \\ 
    \hline
    VQA-T* \cite{yang2021just} & - & - & $C^+$ & DistilBERT & 41.66 & 44.11 & 59.97 & 45.30 & 42.05 & 42.75 & 55.87 & 44.54 \\
    HGA \cite{jiang2020reasoning} & - & - & F, C &BERT & 46.26 & 50.74 & 59.33 & 49.74 & 48.13 & 49.08 & 57.79 & 50.01 \\
    IGV \cite{li2022invariant} & - & - & F, C & BERT & - & - & - & - & 48.56 & 51.67 & 59.64 & 51.34 \\
    HQGA \cite{xiao2021video} & - & - & R, F, C &BERT &48.48 & 51.24 & 61.65 & 51.42 & 49.04 & 52.28 & 59.43 & 51.75 \\
    VQA-T* \cite{yang2021just} & HTVQA69M& 69M & $C^+$ & DistilBERT & 49.60 & 51.49 & 63.19 & 52.32 & 47.89 & 50.02 & 61.87 & 50.83 \\
    P3D-G \cite{cherian2022} &- & - & R, C &BERT & 51.33 & 52.30 & 62.58 & 53.40 & - & -& -& - \\
    ATP \cite{buch2022revisiting} &- & - & F &BERT & 53.1 & 50.2 & 66.8 & 54.3 & - & - & - & -  \\
    \hline
    VGT  & - & - &R, F &BERT & 52.28 & 55.09 & 64.09 & 55.02 & 51.62 & 51.94 & 63.65 & 53.68 \\
    VGT (PT)  & WebVid(WV) & 0.18M &R, F &BERT & 53.43 & 56.39 & \underline{69.50} & 56.89 & 52.78 & 54.54 & \underline{67.26} & 55.70 \\
    CoVGT & - & - &R, F &RoBERTa & \underline{58.80} & \underline{57.44} & 69.37 & \underline{60.01} & {\bf58.53} & \underline{57.02} & 66.83 & \underline{59.42} \\
    CoVGT (PT) & WebVid(WV) & 0.18M &R, F &RoBERTa & {\bf59.69} & {\bf58.00} & {\bf69.88} & {\bf60.73} & \underline{58.00} & {\bf57.96} & {\bf68.40} & {\bf59.69} \\
    \hline
    \end{tabular}
    }
    \label{tab:resnextqa}
    \vspace{-0.5em}
\end{table*}
\subsection{Implementation Details}
We decode the video into frames following \cite{xiao2021video} and sparsely sample $l_v=32$ frames for each video. The frames are distributed into $k=8$ clips whose length $l_c=4$. For each frame, we detect and select the top $n\!=\!10$ regions of high confidence by default (20 for NExT-QA following \cite{xiao2021video} and 5 regions are used in the pretraining-free experiments (see study in Appendix \ref{app:vsp})), using the object detection model from~\cite{anderson2018bottom} which is Faster R-CNN with ResNet-101 backbone pretrained on the Visual Genome dataset~\cite{krishna2017visual}. The frame appearance feature $F_I$ is extracted from ResNet-101 pretrained on ImageNet \cite{russakovsky2015imagenet}. The dimension of the models' hidden states is set to $d=512$, and the default number of graph layers is $U=2$. Besides, the default number of layers and self-attention heads in transformer are $H=1$ and $e=8$ ($e=5$ for edge transformer in DGT) respectively.

For fully-supervised contrastive learning, the negative answers are from two sources comprising the original multiple choices and those sampled from the other questions' correct answers at a probability of 0.3. In open-ended QA, all the other answers in the answer set are treated as the negatives for a given question. For self-supervised contrastive learning, we sample questions from the other samples and treat them as the negative descriptions of the anchor video. In particular, the sampled negatives are from the questions of the same category as the positive question, so as to ensure the hard negatives. The question types are obtained by simple question parsing (see Appendix \ref{app:imp} for details). The trade-off parameter $\lambda$ is set to 1.
%which is searched on the NExT-QA val set and fixed in the other experiments.
We employ Adam \cite{kingma2014adam} optimizer with an initial learning rate of $1\times10^{-5}$ or $5\times10^{-5}$. The learning rate will degenerate following a cosine annealing schedule with respect to the total iterations. The batch size is set to 64, and the maximum epoch varies from 10 to 30 among different datasets. For the pretraining experiment, we download about 0.18M video-text data (less than 10\%) from WebVid2M
\cite{bain2021frozen} and pretrain 2 epochs. More details are presented in the Appendix \ref{app:imp}.

\subsection{State-of-the-Art Comparison}
\label{sec:cpsota}
\subsubsection{Results on NExT-QA}
In Tab. \ref{tab:resnextqa}, we compare CoVGT with some of the lastest graph-based and transformer-based methods on NExT-QA (results per question type are found in Appendix~\ref{app:comp}). The results show that CoVGT significantly surpasses the previous SOTAs on all tasks defined in NExT-QA, especially on the causal (Acc@C) and temporal (Acc@T) reasoning tasks, improving the accuracy on the val and test sets by 5.7\% (\vs ATP \cite{buch2022revisiting}) and 7.7\% (\vs HQGA \cite{xiao2021video}) respectively. Notably, such strong performance does not use external vision-text data for pretraining. The pretrained variants (with 0.18M data) can further increase the accuracy by about 2.0\% for VGT on both the validation and test sets, and 0.7\% for CoVGT on the validation set. The relatively smaller improvement for CoVGT is because pretraining has lost its dominated superiority in answering the descriptive questions (Acc@D). A detailed analysis of the effectiveness of pretraining is presented in Sec.~\ref{sec:pf}.
% The results clearly demonstrate the success of our model.

\textbf{CoVGT \vs VQA-T:}
% In particular, we analyze the comparison with VQA-T~\cite{yang2021just} which also solves VideoQA by contrastive learning instead of classification. 
Compared with VQA-T~\cite{yang2021just} which also solves VideoQA in a contrastive manner but in a supervised fashion analogous to the left term of our Eqn.~\eqref{eq:obj-oe}), our CoVGT wins on several aspects. First, we design DGT for video encoding while VQA-T uses S3D~\cite{xie2018rethinking,miech2020end} (see the benefits of S3D $\rightarrow$ DGT). Second, we design both supervised and self-supervised contrastive objectives which steadily benefit model optimization (see Sec.~\ref{sec:aba}). Third, we encode the question and answer with a single language model, whereas VQA-T encodes question and answer independently with two language models. Our method benefits answer encoding with question as context and reduces model parameters as well (see VGT(DistilBERT)). In addition, such implementation 
%facilitate using more advanced language models (\eg, BERT and RobERTa) which contributes significantly to CoVGT's success (see Sec.~\ref{sec:aba}), and 
permits direct pretraining on user-generated video-text data without the need to generate QA pairs. Finally, VQA-T applies cross-modal transformer (CMTrans) to fuse the video-question pair, whereas we design more light-weight cross-modal interaction module (CM). The results in the 2nd row of Tab.~\ref{tab:comp} indicates that CM has little impact on model performance but reduces model parameters.
\setlength{\tabcolsep}{7pt}
\begin{table}[t!]
    \small
    \centering
    \caption{Detailed comparison between VGT and VQA-T~\cite{Yang_2021_ICCV}. CMTrans: cross-modal transformer.
    }
    \vspace{-0.5em}
    \scalebox{0.8}{
    \begin{tabular}{l|c|ccc|c}
    % \hline\hline
    \multirow{2}*{Models} & \multirow{2}*{Size (M)} & \multicolumn{4}{c}{NExT-QA Val} \\ \cline{3-6}
    ~ & ~ & Acc@C & Acc@T & Acc@D & Acc@All  \\ 
    \hline
    VQA-T \cite{yang2021just} & 600 & 41.66 & 44.11 & 59.97 & 45.30 \\ 
    CMTrans $\rightarrow$ CM &573 & 42.27 & 44.29 & 58.17 & 45.40 \\
    S3D $\rightarrow$ DGT & 641 & 47.53 & 48.08 & 62.42 & 50.02 \\
    \hline
    VGT (DistilBERT) & 346 & 50.71 & 51.67 & 66.41 & 53.46 \\
   
    % \hline
    \end{tabular}
    }
    \vspace{-0.4cm}
    \label{tab:comp}
\end{table}

\textbf{CoVGT \vs HQGA:} HQGA \cite{xiao2022video} constructs graphs on static frames and does not explicitly model the temporal dynamics, whereas we design dynamic visual graphs which exploits the graphs of adjacent frames to regulate the graphs constructed at static frame. Moveover, HQGA extracts language embeddings offline, while we enable online finetuning in an end-to-end fashion. Tab.~\ref{tab:resnextqa} and Tab.~\ref{tab:resqa} show that our method surpasses HQGA on all tasks across different datasets. 
% our graph transformer (\vs pure graph in HQGA) architecture brings better compatibility between the visual and textual feature representations, which may potentially benefits cross-modal interaction and matching. Finally, 
Finally, our stronger results come with more sparse video sampling (see Appendix~\ref{app:vsp}) and without using motion feature. The comparison again points towards the absolute superiority of CoVGT over HQGA.

\textbf{CoVGT \vs P3D-G:} While P3D-G \cite{cherian2022} also applies transformer over visual graphs, the transformer is monolithic and operates over a graph constructed in a pseudo 3D space. First, a single monolithic transformer cannot reflect the local and global nature of video content. Second, to obtain the pseudo 3D graph and register the objects into it, P3D-G needs to transfer 2D RGB frames into RGB-D ones and merge objects globally throughout a whole video. Both processes may accumulate errors from wrong detections and thus jeopardize the performance. In our model, we design local-to-global graph transformer architecture and only link the object within short video clips. Our method is more reasonable in encoding long videos with rich dynamics. 

\textbf{CoVGT \vs ATP:} ATP \cite{buch2022revisiting} focuses on probing key frames from videos for question answering, by using a frozen vision-language model pretrained on image-text data (\ie, CLIP \cite{radford2021learning}). It can well answer questions that invoke frame-level information but may fail to jointly reason over multiple frames to link the atomic things together for compositional and temporal video understanding. In contrast, we have the dynamic graph transformer module to realize this, which we believe, largely contributes to our superior performance. For better analysis, we additionally report results on the ATP-hard subset of NExT-QA validation set. The subset highlights video-level visual-language understanding which is in line with our aim. The results in Tab.~\ref{tab:atphard} show that our CoVGT model significantly surpasses both ATP and its temporal version. 

\setlength{\tabcolsep}{7pt}
\begin{table}[t!]
    \small
    \centering
    \caption{Comparison on ATP-hard subset \cite{buch2022revisiting} of NExT-QA.
    }
    \vspace{-0.5em}
    \scalebox{0.8}{
    \begin{tabular}{l|c|c}
    \multirow{2}*{Methods} & \multicolumn{2}{c}{NExT-QA Val (ATP-hard subset)} \\ \cline{2-3}
    ~ & Acc@C & Acc@T  \\ 
    \hline
    ATP \cite{buch2022revisiting} & 19.6 & 22.6 \\ 
    Temporal[ATP] \cite{buch2022revisiting} & 38.4 & 36.5 \\
    HGA \cite{jiang2020reasoning} & 43.3 & 45.3 \\
    \hline
    CoVGT  & {\bf51.8} & {\bf50.5} \\
    % CoVGT(PT) & \undeline{51.4} & \underline{50.0} \\ 
    % \hline
    \end{tabular}
    }
    \vspace{-0.4cm}
    \label{tab:atphard}
\end{table}

\subsubsection{Results on STAR-QA and Causal-VidQA}
Tab.~\ref{tab:resstar} shows our results on STAR-QA for visual situated reasoning. CoVGT outperforms the previous SOTA (\ie~ClipBERT \cite{lei2021less}) on all the defined four tasks by a clear margin, gaining remarkable improvements of 9.4\% and 7.2\% in mean accuracy with and without pretraining respectively. Again, we find that pretraining steadily boosts performances on all the four tasks, which reveals the strong learning capacity of CoVGT. We also notice that our results surpasses those neuro-symbolic baselines \cite{wu2021star,yi2019clevrer} which relies on additionally functional programs for supervision instead of using only the QA annotations. The observation is in line with the recent work \cite{ding2021attention} which shows the strength of attention over object embedding for complex visual reasoning.
\setlength{\tabcolsep}{8pt}
\begin{table}[t!]
    \small
    \centering
    \caption{Accuracy (\%) comparison on STAR-QA \cite{wu2021star}.  I: Interaction, S: Sequence, P: Prediction, F: Feasibility. M: Mean. Other results are token from \cite{wu2021star}.}
    \vspace{-0.5em}
    \scalebox{0.8}{
    \begin{tabular}{l|cccc|c}
    \hline\hline
    \multirow{3}*{Methods} & \multicolumn{5}{c}{STAR-QA Test} \\ \cline{2-6}
     ~  & Acc@I & Acc@S & Acc@P & Acc@F & \multirow{2}*{Acc@M} \\ 
      ~  & (35.6\%) & (48.4\%) & (9.1\%) & (6.9\%) & ~ \\ 
    \hline
    NS-SR \cite{wu2021star} & 30.88 & 31.76 & 30.23 & 29.73 & 30.65 \\ 
    CLEVRER \cite{yi2019clevrer} & 33.25 & 32.67 & 30.69 & 30.43 & 31.76\\
    \hline
    VisualBERT \cite{li2019visualbert} & 33.59 & 37.16 & 30.95 & 30.84 & 33.14\\
    LGCN \cite{huang2020location} & 39.01 & 37.97 & 28.81 & 26.98 & 33.19 \\
    HCRN \cite{le2020hierarchical} & 39.10 & 38.17 & 28.75 & 27.27 & 33.32 \\
    ClipBERT \cite{fan2019heterogeneous}  & 39.81 & 43.59 & 32.34& 31.42 & 36.79 \\ 
    \hline
    VGT & 42.38 & 47.01 & 41.18 & 39.13 & 42.43 \\
    VGT (PT) & 44.63 & \underline{49.54} & \underline{43.44} & 39.65 & \underline{44.32} \\
    CoVGT & \underline{44.83} & 48.72 & 41.34 & \underline{41.04} & 43.98 \\ 
    CoVGT (PT) & {\bf46.23} & {\bf50.34} & {\bf45.11} & {\bf43.13} & {\bf46.20} \\ \hline
    % Improvements (\%) & \textcolor{red}{+4.8} & \textcolor{red}{+6.0} & \textcolor{red}{+11.1} & \textcolor{red}{+8.0} & \textcolor{red}{+7.5} \\
    % \hline
    \end{tabular}
    }
    \label{tab:resstar}
    \vspace{-0.4cm}
\end{table}

\setlength{\tabcolsep}{10.5pt}
\begin{table*}[t]
    \small
    \centering
    \caption{Accuracy (\%) comparison on Causal-VidQA \cite{li2022representation}. The results of B2A \cite{park2021bridge} are reproduced by \cite{li2022representation} which uses off-the-shelf BERT without finetuning it. (RoI+RoBERTa)'s results are produced by us via filling the Seg\_ID in the texts with the corresponding ground-truth visual object representations (mean-pooled across time for each object.) and sending the combined tokens into RoBERTa for end-to-end finetuning. D: Description, E: Explanation, P: Prediction, C: Counterfactual.
    }
    \vspace{-0.5em}
    \scalebox{0.8}{
    \begin{tabular}{l|c|c|cccccccc|c}
    \hline\hline
    \multirow{3}*{Methods} & \multirow{3}*{Video} & \multirow{3}*{Text} & \multicolumn{9}{c}{Causal-VidQA Test} \\ \cline{4-12}
    ~ & ~ & ~ & \multirow{2}*{Acc@D} & \multirow{2}*{Acc@E} & \multicolumn{3}{c}{Acc@P} & \multicolumn{3}{c|}{Acc@C} & \multirow{2}*{Acc@All} \\
    \cline{6-11}
    ~ &  ~ & ~ & ~ & ~ & Q $\rightarrow$ A & Q $\rightarrow$ R & Q $\rightarrow$ AR & Q $\rightarrow$ A & Q $\rightarrow$ R & Q $\rightarrow$ AR &~ \\ 
    \hline
    B2A \cite{park2021bridge} & F, C & BERT & 66.21 & 62.92 & 48.96 & 50.22 & 31.15 & 53.27 & 56.27 & 35.16 & 49.11 \\
    BlindQA* & - & RoBERTa & 64.65 & 70.03 & 47.39 & 48.39 & 31.87 & 62.94 & 63.12 & 45.26 & 52.95 \\
    RoI+RoBERTa* &  R & RoBERTa & 70.10 & 72.09 & 55.46 & 56.03 & 38.48 & 61.15 & 63.58 & 45.04 & 56.43 \\
    \hline
    CoVGT & R, F &RoBERTa & \underline{73.46} & \underline{74.80} & \underline{58.65} & \underline{56.38} & \underline{39.45} & {\bf66.99} & \underline{64.25} & \underline{48.48} & \underline{59.05} \\
    CoVGT (PT) & R, F &RoBERTa & {\bf74.36} & {\bf75.55} & {\bf60.74} & {\bf60.41} & {\bf43.30} & \underline{65.64} & {\bf64.97} & {\bf50.02} & {\bf60.81} \\
    \hline
    \end{tabular}
    }
    \label{tab:rescausalvid}
    \vspace{-0.4cm}
\end{table*}

our results on Causal-VidQA \cite{li2022representation} are presented in Tab.~\ref{tab:rescausalvid}. The results show that CoVGT remarkably surpasses the previous reported SOTA method (B2A \cite{park2021bridge}) by $\sim$10\% in overall accuracy and beats it on all sub-tasks. We additionally reproduce some stronger baselines (Fine-tuning RoBERTa \cite{liu2019roberta}) for better comparison. The results show that CoVGT still outperforms them by a substantial margin. In addition, the strong performance of a pure language model (\ie~RoBERTa) suggest that the QA contents are biased to text comprehension (also see examples in Appendix~\ref{app:qua_res}). We find that our method performs well on the commonsense reasoning tasks (\ie, Prediction and Counterfactual). We attribute such strong performance to our contrastive learning strategy as well as the exploitation of advanced language models. Moreover, pretraining helps a lot for predicting invisible answers. Additionally, the relatively high accuracy on the separated answer (Q$\rightarrow$A) and reason (Q$\rightarrow$R) but the low joint prediction accuracy (Q$\rightarrow$AR) indicates that model fails to explain the correct answers or wrongly explain the incorrect answers for certain cases. Such failure asks for more future efforts in modelling the consistency between the answers and the corresponding reasons. 
\setlength{\tabcolsep}{3pt}
\begin{table}[t!]
    \small
    \centering
    \caption{Accuracy (\%) comparison. $\dagger$ denotes our newly curated multiple choices by rectifying the redundant options in TGIF-QA-R\cite{peng2021progressive}. We grey out the results reported in \cite{peng2021progressive} regarding these two tasks as the QAs are slightly different.}
    \label{tab:resqa}
    \vspace{-0.5em}
    \scalebox{0.8}{
    \begin{tabular}{l|cc|c|cc|cc}
    \hline\hline
    \multirow{2}*{Methods} & \multicolumn{2}{c|}{Pretrain} & \multirow{2}*{Text} & \multicolumn{2}{c|}{TGIF-QA} & \multicolumn{2}{c}{TGIF-QA-R} \\ \cline{2-3} \cline{5-8}
    ~ & Dataset & Size & ~ & Action & Trans &  Action$^{\dagger}$ & Trans$^{\dagger}$ \\ 
    \hline
    LGCN \cite{huang2020location}& -&-  & GloVe & 74.3 & 81.1  & - & -  \\
    HGA \cite{jiang2020reasoning}& -& -  & GloVe & 75.4 & 81.0  & - & -  \\
    % QueST \cite{jiang2020divide} & -& -  & GloVe & 75.9 & 81.0 & - & -  \\
    HCRN \cite{le2020hierarchical} & -& -  &GloVe & 75.0 & 81.4 & \textcolor[rgb]{0.5,0.5,0.5}{\textit{55.7}} & \textcolor[rgb]{0.5,0.5,0.5}{\textit{63.9}} \\
    % HCRN+ \cite{le2021hierarchical} & -& -  &BERT & 69.8 & 79.8 & -& - \\
    B2A \cite{park2021bridge} & - & -  &GloVe &75.9 & 82.6  & - & -  \\
    HOSTR \cite{dang2021hierarchical} & -& -  &GloVe &75.0 & 83.0  & - & -  \\
    HAIR \cite{liu2021hair} & -& -  &GloVe &77.8 & 82.3 &  - & -   \\
    % MSPAN \cite{guo2021multi} & -& -  & GloVe & 78.4 & 83.3 & - & - \\
    HQGA \cite{xiao2021video} & -& - &BERT &76.9 & 85.6 & - & -   \\
    PGAT \cite{peng2021progressive} & -& -  &GloVe &80.6 & 85.7  & \textcolor[rgb]{0.5,0.5,0.5}{\textit{\underline{58.7}}} & 
    \textcolor[rgb]{0.5,0.5,0.5}{\textit{\underline{65.9}}}  \\
    MASN \cite{seo2021attend} & -& -  &GloVe & 84.4 & 87.4 &- & -  \\
    MHN \cite{peng2022multilevel} & -& - &GloVe &83.5 & 90.8 & - & -   \\
    % DMR \cite{mao2022dynamic} & - & - & GloVe & 84.5 & 90.1 & - & - \\
    \hline
    ClipBERT \cite{lei2021less} & VG, COCO& -  & BERT & 82.8 & 87.8 & - & -  \\
    SiaSRea \cite{yu2021learning} & VG, COCO & -  & BERT & 79.7 & 85.3 & - &-  \\
    % VIOLET \cite{fu2021violet} &  WV & 2.5M  & BERT & 85.5 & 92.1 &- & - \\
    % VIOLET \cite{fu2021violet} &  YT, WV, CC & 185M  & BERT & 92.5 & 95.7 &- & -  \\ 
    MERLOT \cite{zellers2021merlot} & YT, CC & 183M  & BERT & 94.0 & 96.2 & - & -  \\ 
    \hline
    VGT & - & -  &BERT & {\bf95.0} & {\bf97.6} & 59.9 & 70.5 \\
    VGT (PT) & WV & 0.18M  &BERT & 93.1 & \underline{97.2} & 60.5& 71.5 \\
    CoVGT & - & - & RoBERTa & \underline{94.7} & {\bf97.6} & \underline{60.8} & {\bf73.8}  \\
    CoVGT (PT) & WV & 0.18M  &RoBERTa & 91.3 & 96.2 &  {\bf61.0} & \underline{73.2}  \\
    \hline
    \end{tabular}
    }
    \vspace{-0.4cm}
\end{table}

\subsubsection{Results on TGIF-QA and TGIF-QA-R}
In Tab.~\ref{tab:resqa}, we compare our method with previous arts on the TGIF-QA \cite{jang2017tgif} and TGIF-QA-R \cite{peng2021progressive} datasets for repeating action recognition and state transition. The results show that VGT or CoVGT surpasses the previous pretraining-free SOTA results significantly by  $\sim$10\% (VGT \vs MASN \cite{seo2021attend}: 95.0\% \vs 84.4\%) and 7\% (VGT \vs MHN \cite{peng2022multilevel}: 97.6\% \vs 90.8\%) respectively. It even outperforms the pretraining SOTA (\ie~MERLOT \cite{zellers2021merlot}) by about 1.0\%, even though we do not use external data for cross-modal pretraining. On TGIF-QA-R \cite{peng2021progressive} which fixes the answer bias issue in TGIF-QA, CoVGT improves the previous SOTA (PGAT \cite{peng2021progressive}) by about 2\% and 8\% respectively on the repeating action and state transition tasks. The results again demonstrate the strength of our method. 

However, we notice that the improvements of pretraining are unstable on the 4 tasks. We believe that this is because our method has already achieved strong performance without pretraining, and the noises resulted from the pretraining data jeopardize the performances.
A further study by key-word searching the video-text data from WebVid \cite{bain2021frozen} reveals that the pretraining data rarely invoke repeating actions and temporal languages.

\subsubsection{Results on Descriptive QA datasets}
\label{sec:dsp}
While we focus on answering inference-type questions that feature temporal dynamics, we find that CoVGT performs favourably well on the recognition-based VideoQA datasets, \ie, TGIF-FrameQA and MSRVTT-QA. Concretely, our method surpasses the previous object graph-based (pretraining-free) SOTA and shows competitive results to several pretrained Transformer models (shown in Tab.~\ref{tab:resdsp}). Nonetheless, we notice that there is still a clear gap between CoVGT and the SOTA pretrained methods (MERLOT \cite{zellers2021merlot}). The comparison indicates that pretraining with large-scale data is the key to high-ranking results on these datasets, while dense video sampling and relation modelling seem less effective.
\setlength{\tabcolsep}{5pt}
\begin{table}[t!]
    \small
    \centering
    \caption{Accuracy (\%) comparison on descriptive QA datasets.}
    \label{tab:resdsp}
    \vspace{-0.5em}
    \scalebox{0.8}{
    \begin{tabular}{l|cc|c|cc}
    \hline\hline
    \multirow{2}*{Methods} & \multicolumn{2}{c|}{Pretrain} & \multirow{2}*{Text} & TGIF & MSRVTT \\ \cline{2-3}
    ~ & Dataset & Size & ~ & -FrameQA & -QA  \\ 
    \hline
    HOSTR \cite{dang2021hierarchical} & -& -  &GloVe &58.0 & 35.9   \\
    HAIR \cite{liu2021hair} & -& -  &GloVe &60.2 & 36.9    \\
    % MSPAN \cite{guo2021multi} & -& -  & GloVe & 59.7 & 37.8 & 40.3  \\
    MASN \cite{seo2021attend} & -& -  &GloVe & 59.5 & 35.2   \\
    CoMVT \cite{seo2021look} & - & - & BERT & - & 37.3 \\
    % MHN \cite{peng2022multilevel} & -& - &GloVe &58.1 & 38.6   \\
    PGAT \cite{peng2021progressive} & -& -  &GloVe &61.1 & 38.1  \\
    HQGA \cite{xiao2021video} & -& - &BERT &61.3 & 38.6   \\
    % DMR \cite{mao2022dynamic} & - & - & GloVe & 62.5 & 41.6 & -  \\
    % VQA-T \cite{Yang_2021_ICCV} & - & - & DistilBERT & -& 39.6 \\
    \hline
    SSML \cite{amrani2021noise} & HT100M & 100M & BERT & - & 35.1 \\
    CoMVT \cite{seo2021look} & HT100M & 100M & BERT &- & 39.5 \\
    ClipBERT \cite{lei2021less} & VG, COCO& -  & BERT & 60.3 & 37.4   \\
    SiaSRea \cite{yu2021learning} & VG, COCO & -  & BERT & 60.2 & 41.6   \\
    VQA-T \cite{Yang_2021_ICCV} & HT100M & 100M & DistilBERT&- & 40.4 \\
    VQA-T \cite{Yang_2021_ICCV} & HTVQA & 69M & DistilBERT&- & 41.5 \\
    MERLOT \cite{zellers2021merlot} & YT,CC & 185M & BERT & {\bf69.5} & {\bf43.1} \\
    \hline
    VGT & - & -  &BERT & 61.6 & 39.7  \\
    VGT (PT) & WV & 0.18M  &BERT & \underline{61.7} & 39.7 \\
    CoVGT & - & - & RoBERTa & 61.6 & 38.3   \\
    CoVGT  (PT)& WV & 0.18M  &RoBERTa & \underline{61.7} & 40.0   \\
    \hline
    \end{tabular}
    }
    % \vspace{-0.4cm}
\end{table}

\subsection{Model Analysis}
\label{sec:aba}

\subsubsection{Dynamic Graph Transformer}
\textbf{DGT \vs Mean Pooling.} We firstly study the DGT module by substituting it with a simple mean-pooling over the region features; the pooled region features are then interacted with the text features and fed to the global transformer. The ablated model does not capture any spatial and temporal communications between the objects in the local video clips. As shown in the middle part of Tab.~\ref{tab:aba} (w/o DGT), the performances on all datasets drop, especially on those tasks featuring dynamic visual reasoning. For example, the accuracy drops by more than 5\% and 2\% on TGIF-QA Action and Transition datasets respectively. On NExT-QA and STAR-QA, it drops by 2\% and 3\% respectively. This experiment evinces the vital role of DGT.

\textbf{NTrans and ETrans.} We then study the effectiveness of temporal graph transformer in DGT by removing both the node and edge transformers defined in Eqn.~\eqref{equ:node} and \eqref{equ:edge}. Thus, we consider the graphs that are independently constructed at static frames (depicted in Sec.~\ref{sec:vgr}). The results (w/o GTrans) show that this ablation degenerates the overall accuracy by about 1\% on NExT-QA and 2\% on STAR-QA though it performs better than removing the whole DGT module. 

We further study the independent contribution of NTrans and ETrans. We can see that removing any one of them leads to performance drop (VGT \vs w/o NTrans, VGT \vs w/o ETrans). Also, we find that both transformers help improve the performances separately (w/o GTrans \vs w/o NTrans, w/o GTrans \vs w/o ETrans) in most tasks. By comparing the results of w/o NTrans and w/o ETrans, we find that the node transformer contributes relatively more to the results. Such difference is reasonable as the update of the node representations will also update the edges.

Finally, the ablation (w/o $F_I$) in Tab.~\ref{tab:aba} suggests that $F_I$ complements the object graphs well and contributes steadily to the performances across different datasets. 

\setlength{\tabcolsep}{4pt}
\begin{table}[t!]
    \small
    \centering
    \caption{Ablation of architecture designs.
    }
    \label{tab:aba}
    \vspace{-0.5em}
    \scalebox{0.8}{
    \begin{tabular}{l|cc|ccc|c|c}
    % \hline\hline
    \multirow{2}*{Models} & \multicolumn{2}{c|}{TGIF-QA}& \multicolumn{4}{c|}{NExT-QA Val} & STAR Val \\ \cline{2-8}
    ~ & Act & Trans & Acc@C & Acc@T & Acc@D & Acc@All & Acc@M \\ 
    \hline
     VGT & {\bf95.0} & {\bf97.6} & {\bf52.28} & {\bf55.09} & 64.09 & {\bf55.02} & {\bf44.27}  \\ \hline
     w/o DGT & 89.6 & 95.4 &  50.10 & 52.85 & 64.48 & 53.22 & 41.15\\
     w/o TTrans &  94.0 & 97.6 & 50.86 & 53.04 & 64.86 & 53.74 &  42.37 \\
     w/o NTrans & 94.5 & 97.4 & 50.79 & 54.22 & 63.32 & 53.84 & 42.86\\ 
     w/o ETrans & 94.8 & 97.4 & 51.25 & 54.34 & {\bf64.48} & 54.30 & 43.06\\
     w/o $F_I$ & 93.5 & 97.0 & 50.44 & 53.97 & 63.32 & 53.58 & 42.32\\ 
    
    \end{tabular}
    }
    % \vspace{-0.4cm}
\end{table}

\subsubsection{Contrastive Learning}
\label{sec:cst}
\begin{figure}[t!]
  \begin{center}
  \scalebox{1.0}{
    \includegraphics[width=0.48\textwidth]{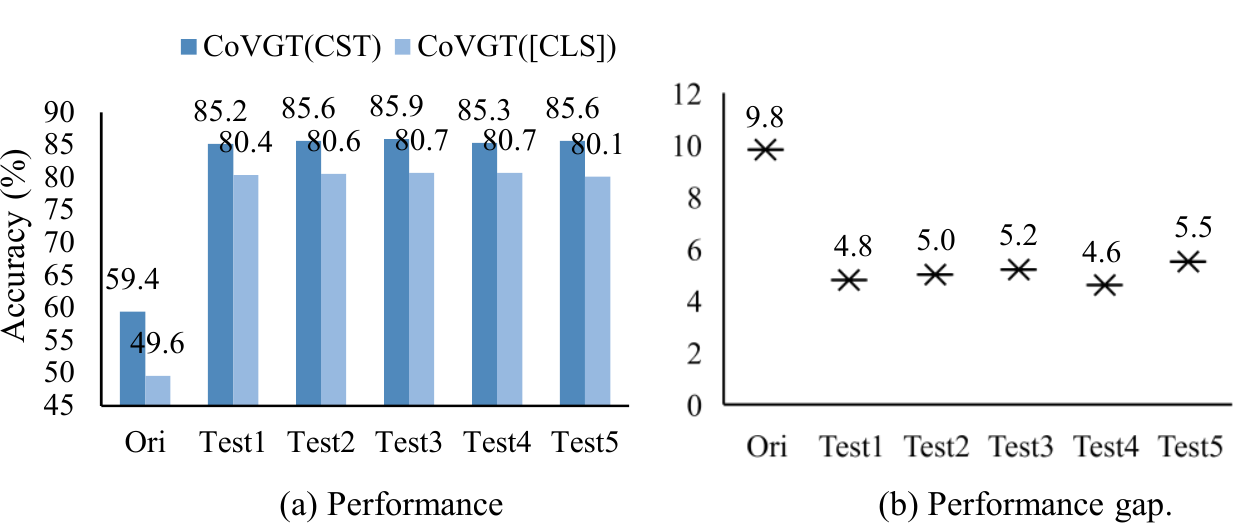}
    }
  \end{center}
  \vspace{-1.0em}
  \caption{Analysis of model generalization to different test sets. We randomly sample 5 times to generate 5 different test sets.}
  \label{fig:newans}
\vspace{-0.4cm}
\end{figure}

\textbf{Contrastive Learning \vs Classification.}
We study a model variant by concatenating the outputs of the DGT module with the token representations from BERT in a way analogous to ClipBERT \cite{lei2021less}. The formed text-video representation sequence is fed to a cross-modal transformer for information fusion. Then, the output of the start token (\eg~\texttt{`[CLS]'} or \texttt{`<s>'}) is fed to a $1$-way classifier for cross-modal matching in multi-choice QA or a $|\mathcal{A}|$-way classifier in open-ended QA. As shown in the 1st row of Tab.~\ref{tab:cst}, this classification model variant ([CLS]) performs poor. A detailed analysis of the performances (see discussion in Appendix \ref{app:cst}.) indicates that the classification layer results in serious over-fitting problem, especially on NExT-QA which has relatively few training data while the QA contents are complex and diverse. We additionally compare the two kinds of learning strategy on new answer distributions. To this end, we keep unchanged the test questions and their correct answers, but replace the negative answers with randomly sampled answers. Fig.~\ref{fig:newans}(a) suggests that both the classification and contrastively learned models show generalization capability on the newly curated test data. Nonetheless, we find that the performance gap in Fig.~\ref{fig:newans}(b) becomes smaller than it on the original test set. Such difference indicates that the classification model is not good at disambiguating the hard negatives, because the original negative answers are carefully curated to be much more challenging than that of random sampling. These experiments demonstrate the superiority of solving multi-choice QA by contrastive learning over classification. 
\setlength{\tabcolsep}{2pt}
\begin{table}[t!]
    \small
    \centering
    \caption{Study of contrastive learning. RBT: RoBERTa.}
    \label{tab:cst}
    \vspace{-0.5em}
    \scalebox{0.8}{
        \begin{tabular}{l|ccc|cccc}
        Models & $\mathcal{L}_{vqa}$ & $\mathcal{L}_{vq}$ & RBT & NExT-QA & STAR-QA & TGIF-R-Trans$^{\dagger}$ & TGIF-FQA \\ 
        \hline
        [CLS] & ~ & ~ & ~ & 45.82 & 41.91 & 65.9 & 56.9 \\
        VGT & $\surd$& ~ & ~ & 55.02 & 44.27 & 70.5 & 61.6 \\
        ~ & $\surd$ & $\surd$ & ~ &  57.15 & 45.84 & 71.6 & 61.3\\ 
        ~ & $\surd$ &~ & $\surd$ &  58.45 & 42.44 & 72.0 & 60.6 \\
        CoVGT & $\surd$ & $\surd$ & $\surd$ & {\bf60.01} & {\bf45.97} & {\bf73.8} & {\bf61.6} \\
        \end{tabular}
    }
    \vspace{-0.5cm}
\end{table}

\begin{figure}[t!]
  \begin{center}
    \includegraphics[width=0.48\textwidth]{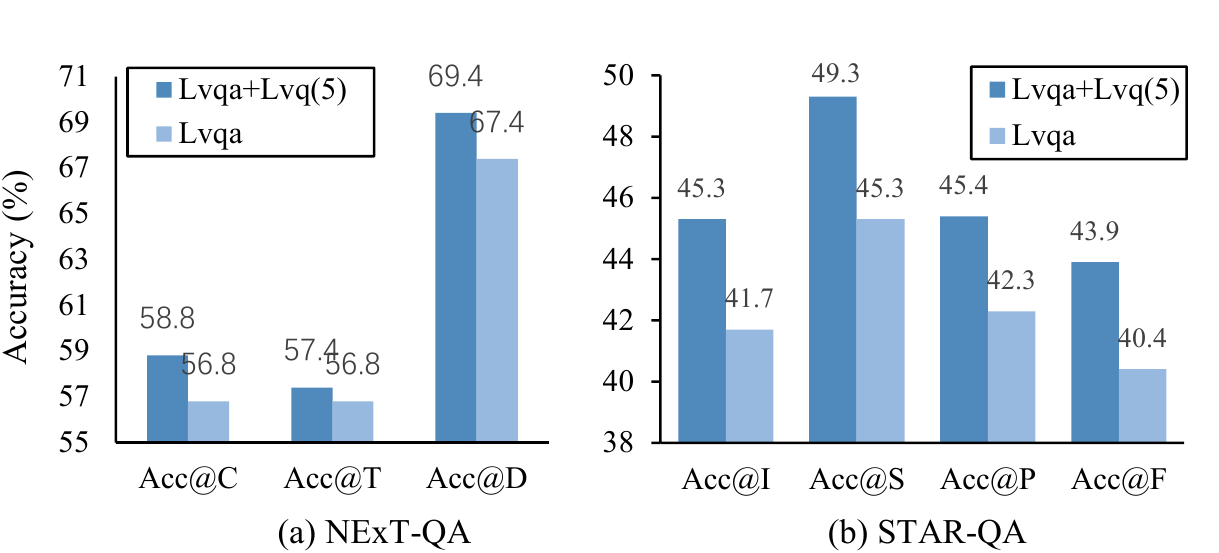}
  \end{center}
  \vspace{-1em}
  \caption{Comparison of our methods with and without $\mathcal{L}_{vq}$ on specific tasks. The results are based on the language encoder RoBERTa. $\mathcal{L}_{vq}(5)$ means 5 negative questions.}
  \label{fig:cst}
 % \vspace{-0.4cm}
\end{figure}

\begin{figure*}[t!]
  \begin{center}
    \includegraphics[width=0.9\textwidth]{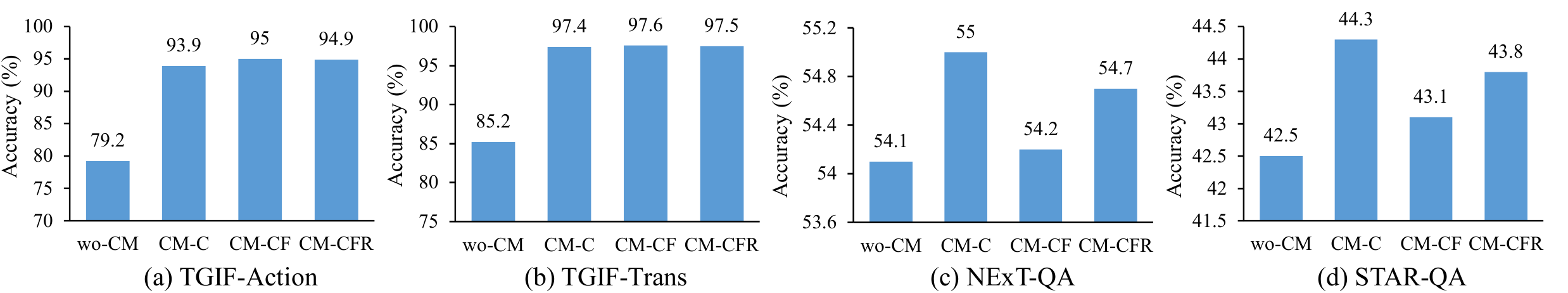}
  \end{center}
  \vspace{-0.4cm}
  \caption{Investigation of cross-modal interaction.}
  \label{fig:cm}
  \vspace{-0.4cm}
\end{figure*}
\textbf{Supervised and Self-supervised Contrastive Learning.}
Tab.~\ref{tab:cst} shows that the self-supervised contrastive objective function ($\mathcal{L}_{vq}$) coordinates well with the supervised one ($\mathcal{L}_{vqa}$). It improves the performances on different datasets with different language encoders (BERT \cite{devlin2018bert} by default and RoBERTa \cite{liu2019roberta}) by a clear margin. The detailed results on NExT-QA and STAR-QA (see Fig.~\ref{fig:cst}) show that the improvements are also stable across different tasks. Concretely, the improvements on the causal and temporal reasoning tasks of NExT-QA are more remarkable than that of the descriptive task. Our explanation is that the descriptive questions are relatively simple and involve few visual concepts for cross-modal correspondence learning, \eg, \texttt{`what/where is this happening'}. Therefore, the self-supervised objective helps little in that case. All of the questions in STAR-QA are populated from scene graph annotations and challenge visual relation reasoning, so the improvements are significant on all tasks. In particular, we find that the improvements on the rare questions (\ie, questions in the Feasibility groups) are quite impressive though they have few training samples. The improvements, along with our observation of the train and validation accuracy during training, suggest that $\mathcal{L}_{vq}$ can alleviate the over-fitting problem and hence enhance the model's generalization capability.

\textbf{Negative Sample Mining.}
We study the number of negative questions in self-supervised contrastive learning ($\mathcal{L}_{vq}$) based on NExT-QA. For supervised learning ($\mathcal{L}_{vqa}$), the number of negative answers is kept the same as the original multiple choices. Tab.~\ref{tab:abaneg} shows that the number of negative samples has relatively little impact (\eg, less than 1.0\%) on the accuracy. In addition, learning with different number of negative samples can steadily improve over the baseline that does not use video-question correspondence as auxiliary supervision. Intriguingly, we find that the best result is achieved with 4 negative samples. The number is in line with the number of native answers in multi-choice QA. 
\setlength{\tabcolsep}{5pt}
\begin{table}[t!]
    \small
    \centering
    \caption{Study the number of negative questions in $\mathcal{L}_{vq}$.}
    \label{tab:abaneg}
    \vspace{-0.5em}
    \scalebox{0.8}{
        \begin{tabular}{l|ccccc}
        % \hline
        \#Negative samples & 0 & 4 & 9 & 14 & 19 \\
        \hline
        Accuracy (\%) & 58.45 & {\bf60.01} & 59.85 & 59.67 & 59.81 \\
        \end{tabular}
    }
    \vspace{-0.4cm}
\end{table}

In Tab.~\ref{tab:abahneg}, we study the effectiveness of hard negative sampling. Our observations are as follows: 1) The hard negative sampling can steadily improve the performances over a random selection (row 2 and 3 \vs row 1). 2) Our method of parsing the questions to obtain the question types works well in experiments. It can achieve equivalent results to the method of using the ground-truth question types (row 2 \vs row 3). 3) We surprisingly find that a random selection of the negatives can also contribute to the performance (row 1). Such observation reveals the strength of our contrastive learning.
\setlength{\tabcolsep}{5pt}
\begin{table}[t!]
    \small
    \centering
    \caption{Study hard negative samples. Random: randomly sample the negatives. PQ: our method by parsing the questions. GT: use the ground-truth type annotations.}
    \label{tab:abahneg}
    \vspace{-0.5em}
    \scalebox{0.8}{
        \begin{tabular}{l|ccc}
        % \hline
        Methods & NExT-QA & STAR-QA & Causal-VidQA \\
        \hline
       Random & 59.35 & 45.40 & 59.44 \\
       Type (PQ) & {\bf60.01} & 45.97 & 59.67 \\
       Type (GT) & 59.93 & {\bf45.98} & {\bf59.97} \\
       \end{tabular}
       }
    % \vspace{-0.4cm}
\end{table}

\subsubsection{Cross-modal Interaction}
In Fig.~\ref{fig:cm}, we investigate several implementation variants of the cross-modal interaction module as depicted in Sec.~\ref{sec:cm}. The results suggest that it is better to integrate textual information at both the frame- and clip-level outputs (\ie, Eqn.~\eqref{equ:ff} and~\eqref{equ:fc} respectively) for TGIF Action and Transition datasets, while a simple interaction at the clip-level (by default) brings the optimal results on other datasets. 
% Besides, the region-level interaction seems to be redundant for all experiments. We speculate that the frame- and clip-level representations have already absorbed the region-level information. 
Based on these observations, we operate the cross-modal interaction module at both the frame- and clip-level outputs for TGIF Action and Transition, and keep the default implementation (interaction at the clip-level) for other datasets.
% so as to gain both accuracy and efficiency. 

Compared with the baselines without cross-modal interaction, all three kinds of interactions help to improve the performance. 
% This is because cross-modal interaction can help identify the relevant visual information from videos for question answering. 
We notice that the improvement on TGIF is more than 10\%. Our explanation is as follows: GIFs are trimmed short videos that only contain the QA-related visual content. The simple video content greatly eases the challenge in spatial-temporal grounding of the positive answer, especially when most of the negative answers do not appear in the short GIFs and can be well distinguished by advanced language models. Therefore, the cross-modal interaction performs more effectively on this dataset. While the video clips in STAR-QA are also trimmed, its multiple choices are carefully curated to include at least one distractor answer that appears in the same video clip. Thus, the improvements are relatively smaller.

\subsection{Pretraining and Finetuning}
\label{sec:pf}

\setlength{\tabcolsep}{8pt}
\begin{table}[t!]
    \small
    \centering
    \caption{Contrastive pretraining with a different number of negative samples. Results are reported on NExT-QA test set.}
    \label{tab:ptneg}
    \vspace{-0.5em}
    \scalebox{0.8}{
        \begin{tabular}{l|cccccc}
        % \hline
        \#Negative & 4 & 15 & 31 & 63 & 127 & 255 \\
        \hline
        Zero-shot & 31.9 & 33.1 & {\bf34.5} & 34.2 & 34.5 & 34.3 \\
        Fine-tune & 55.1 & 55.3 & {\bf55.7} & {\bf55.7} & 55.5 & 55.2 \\
        \end{tabular}
    }
    \vspace{-0.4cm}
\end{table}
% We study the effects of \emph{pretrain and finetune} based on VGT.
\textbf{Number of Negative Samples.}
In Tab.~\ref{tab:ptneg}, we study pretraining with a different number of negative samples based on VGT. We find that:
% \begin{enumerate}
1) the number of negative samples mostly affects zero-shot QA performances; when finetuning on the target dataset, the differences become smaller (\eg~$<1\%$); and
2) relatively more negative samples give rise to better results, and the best result is achieved at 31 and 63. Note that the number of negative answers in the target datasets is much smaller (\eg~4 or 5).
\begin{figure}[t!]
  \begin{center}
    \includegraphics[width=0.45\textwidth]{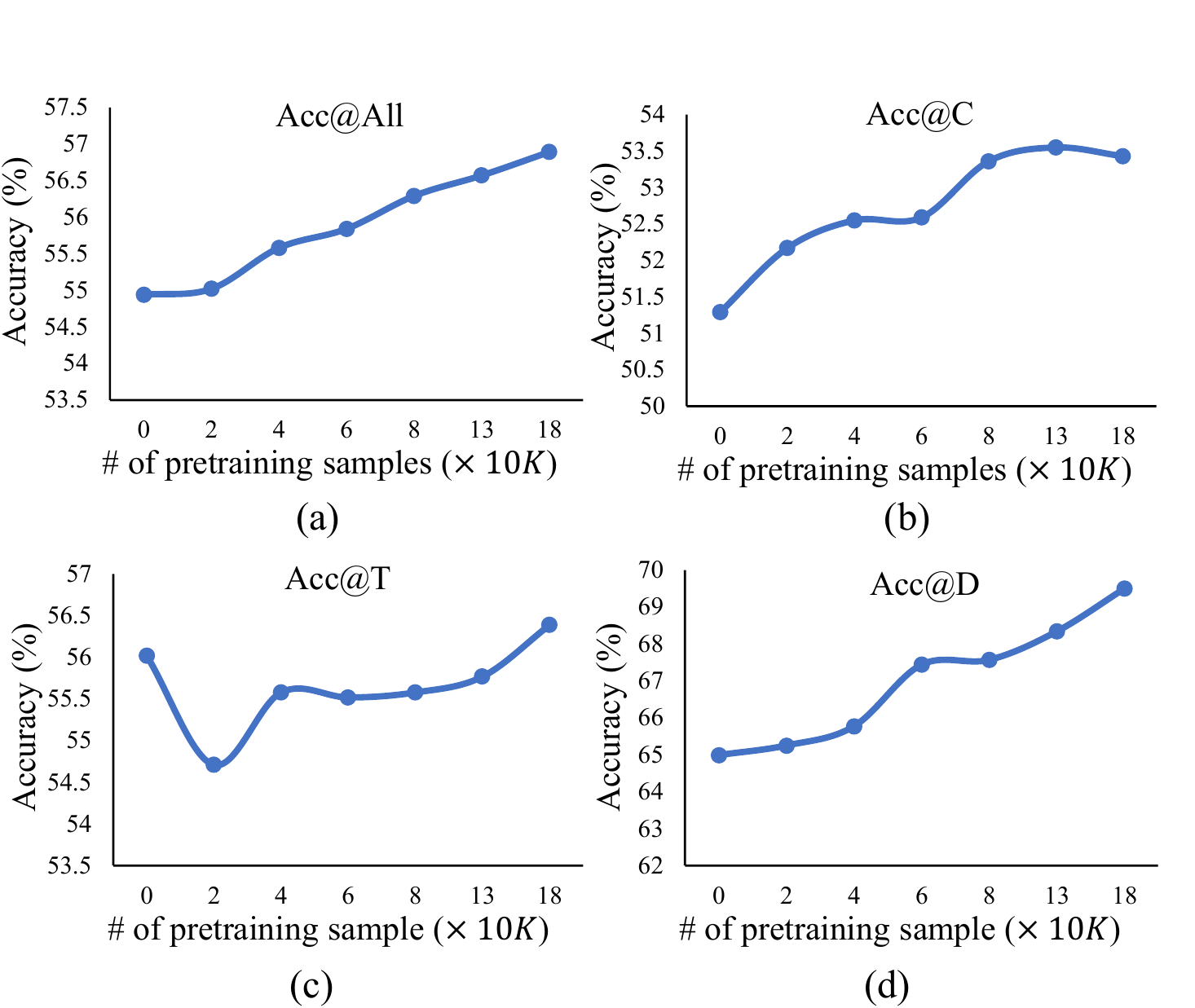}
  \end{center}
  \vspace{-1em}
  \caption{Pretraining with different amount of data. Results are reported on NExT-QA validation set.}
  \label{fig:pts}
  % \vspace{-0.4cm}
\end{figure}

\textbf{Amount of Pretraining Data.}
In Fig.~\ref{fig:pts} we study VGT's performances with different size of pretraining data. Generally, we can see that there is a clear tendency of performance improvements for the overall accuracy (Acc@All) when more data are available. A more detailed analysis shows that these improvements mostly come from a stronger performance in answering causal (Acc@C) and descriptive (Acc@D) questions. It seems that to answer the descriptive questions well, we just need more data for pretraining. However, for answering temporal questions, it demands relatively more data to yield positive effect, or otherwise pretraining helps little and even hurts the performance. This could be because our pretraining data (\ie~WebVid) rarely invokes temporal descriptions and the proxy tasks are not tailored for temporal reasoning. 
Finally, answering casual questions is still the most challenging tasks since the accuracy on causal questions is the lowest among the three tasks categorised in NExT-QA. The observations advocate more future efforts in exploring pretraining to better handle temporal and causal visual reasoning problems.

\setlength{\tabcolsep}{5pt}
\begin{table}[t!]
    \small
    \centering
    \caption{Study of finetuning the pretrained (PT) weights.}
    \label{tab:pt}
    \vspace{-0.5em}
    \scalebox{0.8}{
        \begin{tabular}{l|cccc|ccc|c}
         Models & $\mathcal{L}_{vqa}$ & $\mathcal{L}_{vq}$ & $\mathcal{L}_{\text{MLM}}$ & PT & Acc@C & Acc@T & Acc@D & Acc@All \\ 
        \hline
        \multirow{4}*{VGT}
        &$\surd$ & ~ & ~ & ~ & 52.28 & 55.09 & 64.09 & 55.02 \\
        ~&$\surd$ & ~& $\surd$ & ~ & 49.41 & 54.59 & 64.74 & 53.46 \\
        ~&$\surd$ &~ & ~ & $\surd$& 52.28 & 55.77 & 69.11 & 56.02 \\
        ~&$\surd$ & ~ & $\surd$ & $\surd$ &  {\bf53.43} & {\bf56.39} & {\bf69.50} & {\bf56.89} \\ \hline
        \multirow{2}*{CoVGT}
        & $\surd$ & $\surd$ &~ & $\surd$ &  58.42 & {\bf58.25} & 67.82 & 59.83 \\
        ~& $\surd$ & $\surd$ & $\surd$ & $\surd$ & {\bf59.69} & 58.00 & {\bf69.88} & {\bf60.73} \\
        \end{tabular}
    }
    \vspace{-0.4cm}
\end{table}

\begin{figure*}[t!]
  \begin{center}
    \includegraphics[width=1.0\textwidth]{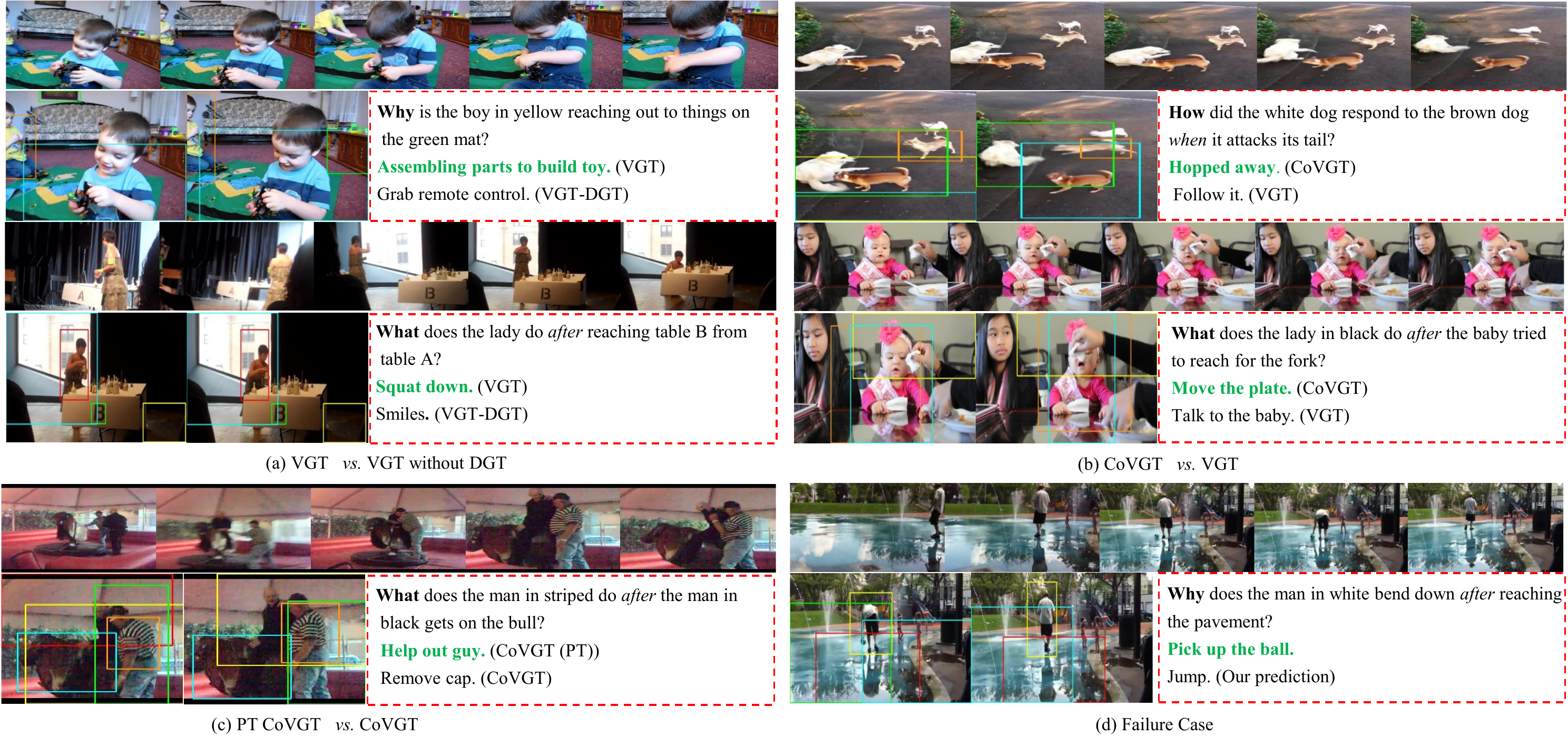}
  \end{center}
  \vspace{-0.2cm}
  \caption{Visualization of typical predictions on NExT-QA \cite{xiao2021next}. The ground-truth answers are highlighted in green. }
  \label{fig:vis}
  \vspace{-0.4cm}
\end{figure*}
\textbf{Fine-tuning.}
\label{sec:ft}
In Tab.~\ref{tab:pt}, we study whether finetuning with masked language modelling (MLM) will also help QA performance. The results show that adding the MLM objective for finetuning the pretrained weights can steadily improve the performance for both VGT and CoVGT. Nevertheless, when learning the model from scratch with MLM, such advantage disappears as there is a performance drop on causal and temporal reasoning tasks. The observations indicate that learning with masked language modelling on causal and temporal reasoning tasks demands more data. We also analyze finetuning by adapting the multi-choice QA-pairs to descriptions (see Appendix \ref{app:adapt}), so as to reduce the task gap between pretrain and finetune. Our conclusion is that the adaptation method does help improve the performance but the improvement is not stable, and thus we do not use it.

\subsection{Model Efficiency}
\setlength{\tabcolsep}{5pt}
\begin{table}[t!]
    \small
    \centering
    \caption{Comparison of memory and time based on NExT-QA \cite{xiao2021next}. (2m$\times$8: 2 minutes per epoch and 8 epochs in total.) 
    }
    \vspace{-0.5em}
    \scalebox{0.8}{
    \begin{tabular}{l|c|c|cc|cc}
    % \hline\hline
    \multirow{2}*{Models} & \multirow{2}*{Acc(\%)} &\multirow{2}*{\#Params} & \multicolumn{2}{c|}{GPU Memory} & \multicolumn{2}{c}{Time} \\ 
    \cline{4-7}
    ~ & ~ & ~ & Train & Infer & Train & GFLOPs \\ 
    \hline
    VQA-T \cite{yang2021just} &45.30 & 156.5M & 5.6G & 2.6G & 2m$\times$8 & 2.5 \\ 
    VGT (DistilBERT) & 53.46 & 90.5M & 10.0G & 3.5G & 5m$\times$7 & 3.9 \\ 
    VGT (BERT) & 55.02 & 133.7M & 16.2G & 3.9G & 7m$\times$5 & 7.1 \\ 
    CoVGT (RoBERTa) & 60.01 & 148.9M & 19.6G & 4.0G & 7m$\times$7 & 12.7 \\ 
    % CoVGT & 59.49 & 148.9M & 19.6G & 4.0G & 7m$\times$7 & 12.7 \\ 
    % \hline
    \end{tabular}
    }
    \label{tab:tm}
    \vspace{-0.4cm}
\end{table}

We compare CoVGT with VQA-T in Tab.~\ref{tab:tm} for better understanding of model efficiency. Experiments are conducted on 1 Tesla V100 GPU with batch size 64 (GFLOPs are based on 1 example).
i) \textbf{{Memory:}} CoVGT has comparable training parameters (148.9M \vs 156.5M) and model size with VQA-T (568M \vs 600M). The RoBERTa encoder in CoVGT takes large portion of the parameters, the vision part is lightweight with only 24M parameters. CoVGT needs more GPU memory for training. Yet, the memory for inference is fairly small and close to that of VQA-T. 
ii) \textbf{{Time:}} CoVGT's running speed is lower than VQA-T. The results reveal that the lower speed is mainly resulted from the language encoder. 
However, CoVGT converges much faster and needs much fewer epochs (total FLOPs) to get results superior to VQA-T. For example, on NExT-QA, CoVGT's accuracy at epoch 1 is 50.76\% and 55.1\% at epoch 2, which already significantly surpass VQA-T's best result (45.30\%) achieved at epoch 8. Also, CoVGT's result without pretraining can surpasses that of VQA-T pretrained with million-scale data. In that sense, VGT needs much fewer total FLOPs than VQA-T and other similar pretrained models to achieve better results of visual reasoning. 

\subsection{Qualitative Analysis}
\label{sec:qua}
\textbf{VGT \vs VGT-DGT}.
In Fig.~\ref{fig:vis}(a), the ablated model wrongly answers the 1st question with an atomic action \texttt{`grab'} without DGT to weave together a series of \emph{boy}-\emph{toy} interactions (\eg, \texttt{`tough, grab, plug , ...'}) to achieve \texttt{`assemble'}. For the 2nd question, prediction like \texttt{`squat down'} requires the model to capture the object's temporal state changes, otherwise the model tends to predict the static action \texttt{`smiles'} for answer. The examples demonstrate the effectiveness of DGT in modelling the compositions and temporal-dynamics.

\textbf{CoVGT \vs VGT}.
Fig.~\ref{fig:vis}(b) reveals that CoVGT is able to predict the answers that have never been the correct answers during training (\eg~\texttt{`hopped away'} and ~\texttt{`move the plate'}). Yet, the words \texttt{`hop'} and \texttt{`plate'} can be found in the training questions. Such generalization capability of predicting unseen/rare answers mainly thanks to CoVGT's self-contrastive learning strategy between the relevant and irrelevant questions.

\textbf{PT CoVGT \vs CoVGT}.
Fig.~\ref{fig:vis}(c) suggests that pretraining (PT) helps the model predict the abstract answers, \eg~\texttt{`help'}. This is understandable since the abstract words often correspond to diverse video contents, and thus demand more data for learning.

\textbf{Failure Cases}.
The example in Fig.~\ref{fig:vis}(d) shows a failure case where our model wrongly answers the question with \texttt{`jump'}. The visualization of the detection results suggest that the detection model fail to detect the small object, \eg~\texttt{`ball'}. The case indicates that understanding of the fine-grained object interaction is still challenging and needs more future efforts.
%--------------------------------------------------------
\subsection{Limitations and Opportunities}
% While we have explored a novel way to perform VideoQA and achieved strong results across broad-ranging datasets. 
We discuss several limitations and leave them as future efforts. First, we pre-sample and -extract video features offline, which may leave out some key frames and objects that are important for question answering (see our analysis in Sec.~\ref{sec:qua}). We believe that an online approach which can take into account more video contents (\eg, at different training iterations) could be helpful for performance improvement. Second, our method %more or less 
benefits from large-scale language models; accordingly it requires more memory and time for inference %, yet also suffers from their relatively lower efficiency 
(see Tab.~\ref{tab:tm}). Therefore, study of light-weight models that are capable of complex video reasoning is a promising direction. Finally, our improvement on open-ended QA (common setting for recognition-based QA) is smaller than that of multi-choice QA (common setting for video reasoning) (Tab.~\ref{tab:resdsp} \vs other tables in Sec.~\ref{sec:cpsota}), though we have shown its effectiveness for both tasks. Thus, it is interesting to explore more effective approaches for open-ended QA. Our findings in Tab.~\ref{tab:resdsp} and Fig.~\ref{fig:pts}(d) suggest that data is essential whereas relation modeling seems less effective. Finally, our experiments are tied to the VideoQA task, it would be interesting to examine our DGT video encoder in other video understanding tasks.

\section{Conclusion}
\label{sec:con}
This paper introduced a contrastively learned video graph transformer (CoVGT) model for VideoQA in a totally contrastive manner. The model mainly includes: 1) a dynamic visual graph transformer module along with a local-to-global hierarchical architecture for video encoding, 2) separate visual and textual transformers along with a light-weight cross-modal interaction module for cross-modal information encoding, communication and comparison, and 3) joint fully-supervised and self-supervised contrastive objective function for parameter optimization. To validate CoVGT's effectiveness and the contribution of each component, 
we conducted extensive experiments on a wide range of benchmarks that challenge the various aspects of cross-modal video understanding. The results show that CoVGT can remarkably improve the previous SOTA results on video reasoning tasks and obtain competitive results on descriptive QA tasks.
We further demonstrated that our model can also benefit from cross-modal pretraining. Our success suggests that with careful engineering of architectures and learning strategy, one can significantly lower the burden of handling large-scale video data for pretraining, yet achieve comparable or superior performance.

\ifCLASSOPTIONcompsoc
\section*{Acknowledgments}
\else
\section*{Acknowledgment}
\fi
This research is supported by the Sea-NExT Joint Lab. The research is also supported by the National Natural Science Foundation of China under grant 61932009.

\ifCLASSOPTIONcaptionsoff
  \newpage
\fi

{
\bibliographystyle{IEEEtran}
\bibliography{egbib}
}

\begin{IEEEbiography}[{\includegraphics[width=1in,height=1.25in,clip,keepaspectratio]{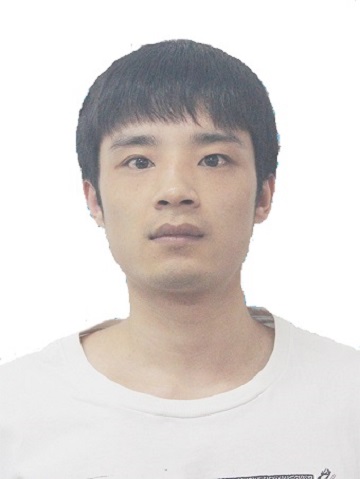}}]{Junbin Xiao} has completed his PhD and now serves as a Research Fellow at the Department of Computer Science (CS), National University of Singapore (NUS). Before that, he received his M.S.Eng degree from the Institute of Computing Technology (ICT), Chinese Academy of Sciences (CAS) and B.E degree from Sichuan University (SCU). His research focuses on visual relation oriented VideoQA. He has published relevant papers in the top-tier conferences: CVPR, ECCV, AAAI, ACM MM and EMNLP. He also serves as reviewer for: CVPR, ICCV, ECCV, AAAI, TMM, ToMM and TNNLS. He co-organized and served as program committee member for the video relation understanding challenge on MM'19 and MM'20.
\end{IEEEbiography}

\begin{IEEEbiography}[{\includegraphics[width=1in,height=1.25in,clip,keepaspectratio]{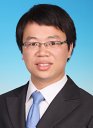}}]{Pan Zhou} is currently a senior Research Scientist at Sea AI Lab (SAIL) of Sea group. Before that, he worked in Salesforce as a research scientist. He completed his Ph.D. at National University of Singapore (NUS) and Master at Peking University. His research interests include deep learning theory and applications, noncovex/convex optimization. He has published papers in ICLR, ICML, NeurIPS, CVPR, ICCV, ECCV, AAAI, IJCAI and journals: TPAMI, TIP. He serves as reviewer for top conferences: ICML, NeurIPS, CVPR, ICCV, AAAI and journals: TPAMI, IJCV, TIP, TNNLS and TCSVT. He is awarded the Microsoft Research Asia Fellowship.
\end{IEEEbiography}

\begin{IEEEbiography}[{\includegraphics[width=1in,height=1.25in,clip,keepaspectratio]{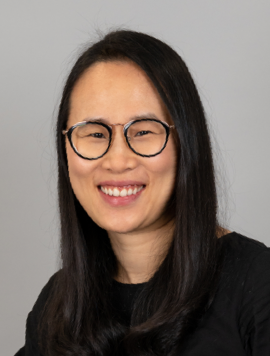}}]{Angela Yao} is currently an assistant professor and leads the Computer Vision and Machine Learning (CVML) group at National University of Singapore. She received her Ph.D degree from ETH and BASc degree from University of Toronto, Canada. Her research interests include video understanding, 3D pose estimation, activity recognition, sequence and time series modelling. Her research lies at the intersection of computer vision, machine learning, and human-computer interaction, with a particular emphasis on developing models and algorithms that can understand human actions and interactions in video data. She serves as Area Chair and Reviewer for ICLR, ICML, NeurIPS, CVPR, ICCV, IJCAI and Program Chair for ICCV, ECCV and 3DGV. She once led the Visual Computing Group at the University of Bonn, Germany and co-founded a startup on smart parking in Zurich, Switzerland.
\end{IEEEbiography}

\begin{IEEEbiography}[{\includegraphics[width=1in,height=1.25in,clip]{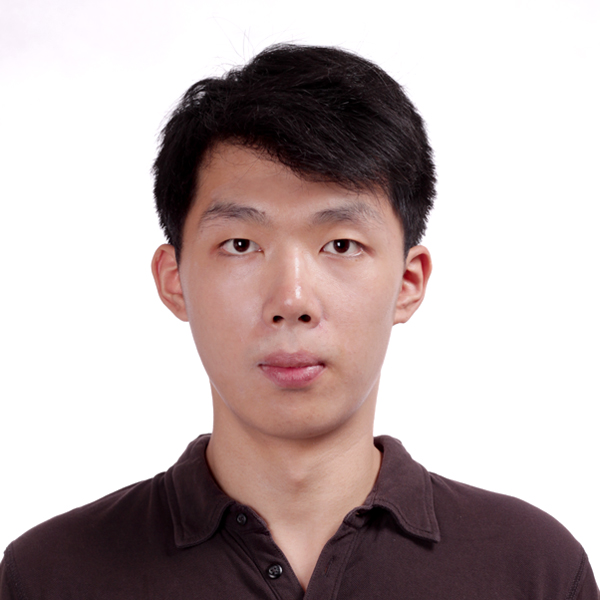}}]{Yicong Li} is currently a Ph.D. candidate at the Institute of Data Science, National University of Singapore. Prior to that, he received a bachelor's degree and a Master's degree from Huazhong University of Science and Technology and Columbia University, respectively. In 2020, he started his Ph.D. study at the National University of Singapore with a research focus on multi-modal learning. His publications include some top conferences such as CVPR, ACM MM, AAAI and EMNLP. He serves as reviewer for CVPR.
\end{IEEEbiography}

\begin{IEEEbiography}[{\includegraphics[width=1in,height=1.25in,clip]{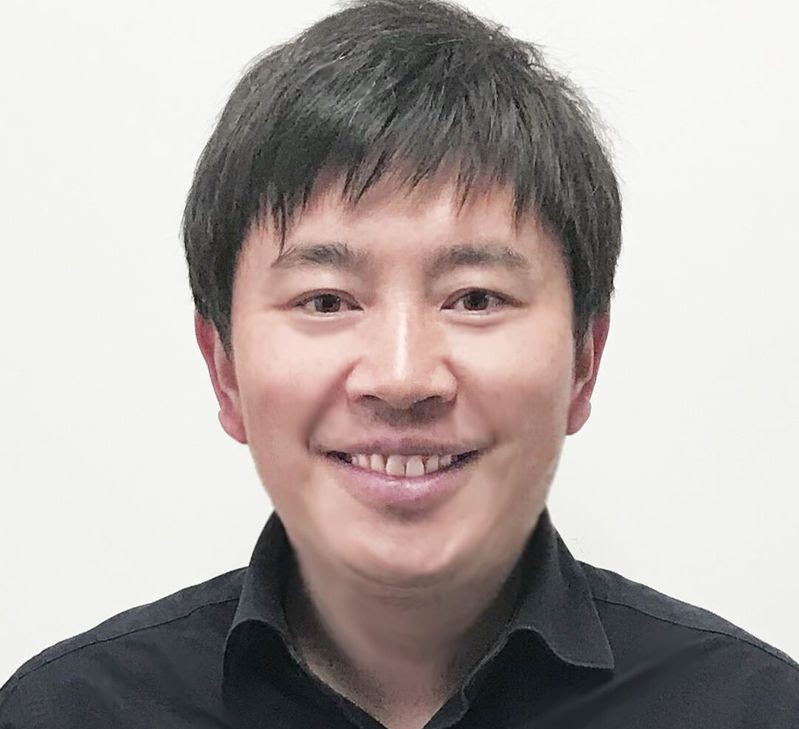}}]{Richang Hong} is the Professor and Executive Dean
at School of Computer and Informatics, Hefei University of Technology, Hefei, China. He received his Ph.D. degree from the University of Science and Technology of China (USTC). He worked as a research fellow at National University of Singapore (NUS). His current research interests include multimedia, language and vision and social media. He has authored over 200 journal and conference papers in these areas and the Google Scholar citations for those papers is more than 16000. He served as editor of the IEEE TCSVT, IEEE TMM, IEEE TCSS, IEEE TBD, ACM TOMM, NPL and the guest editors of several international journals, a steering committee member of MMM (international conference on multimedia modeling) conference series since 2019, and the technical program chairs of PCM'2018, etc. He also served as area chairs of ACM Multimedia, SIGIR etc. since 2016 and a technical program committee member of over 20 prestigious international conferences, and a reviewer of over 20 prestigious international journals. He is a recipient of the Best Paper Award in ACM Multimedia 2010, Best Paper Award in ACM ICMR 2015 and Best Paper Honorable Mention Award of IEEE trans. Multimedia 2015. Dr. Hong joined CCF, CSIG, CAAI, IEEE and ACM. He is the deputy director of multimedia technical committee of CSIG and the secretary of the ACM SIGMM China Chapter.
\end{IEEEbiography}

\begin{IEEEbiography}[{\includegraphics[width=1in,height=1.25in,clip]{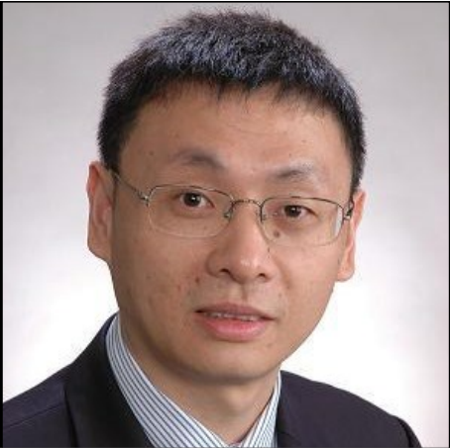}}]{Shuicheng Yan} (Fellow, IEEE) is currently a visiting professor at BAAI, Beijing, China. Previously, he was the director of Sea AI Lab (SAIL) and Group Chief Scientist of Sea. He is an Fellow of Academy of Engineering, Singapore, IEEE Fellow, ACM
Fellow, AAAI Fellow and IAPR Fellow. His research areas include computer vision, machine learning and multimedia analysis. Till now, he has published over 600
papers in top international journals and conferences, with Google Scholar Citation over 90,000 times and H-index 135. He had been among
“Thomson Reuters Highly Cited Researchers” in 2014, 2015, 2016, 2018, 2019. Dr. Yan’s team has received winner or honorable-mention prizes for 10 times of two core competitions, Pascal VOC and ImageNet (ILSVRC), which are deemed as “World Cup” in the computer vision community. Also his team won over 10 best paper or best student paper prizes and especially, a grand slam in ACM MM, the top conference in multimedia, including Best Paper Award, Best Student Paper Award and Best Demo Award.
\end{IEEEbiography}

\begin{IEEEbiography}[{\includegraphics[width=1in,height=1.25in,clip,keepaspectratio]{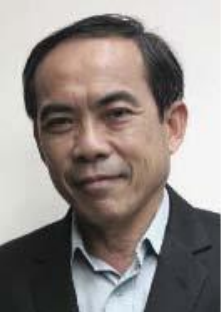}}]{Tat-Seng Chua} is the KITHCT Chair Professor at the School of Computing, National University of Singapore (NUS). He is also the Distinguished Visiting Professor of Tsinghua University, the Visiting Pao Yue-Kong Chair Professor of Zhejiang University, and the Distinguished Visiting Professor of Sichuan University. Dr. Chua was the Founding Dean of the School of Computing from 1998-2000. His main research interests include unstructured data analytics, video analytics, conversational search and recommendation, and robust and trustable AI. He is the Co-Director of NExT, a joint research Center between NUS and Tsinghua University, and Sea-NExT, a joint Lab between Sea Group and NExT.
Dr. Chua is the recipient of the 2015 ACM SIGMM Achievements Award, and the winner of the 2022 NUS Research Recognition Award. He is the Chair of steering committee of Multimedia Modeling (MMM) conference series, and ACM International Conference on Multimedia Retrieval (ICMR) (2015-2018). He is the General Co-Chair of ACM Multimedia 2005, ACM SIGIR 2008, ACM Web Science 2015, ACM MM-Asia 2020, and the upcoming ACM conferences on WSDM 2023 and TheWebConf 2024. He serves in the editorial boards of three international journals. Dr. Chua is the co-Founder of two technology startup companies in Singapore. He holds a PhD from the University of Leeds, UK.
\end{IEEEbiography}

\clearpage
\appendices
% \section{Experiments}
This appendix gives additional introduction to the paper "Contrastive Video Question Answering via Video Graph Transformer". It includes three major parts: A. more information of the experimented datasets, B. the implementation details, and C. more result discussion. 

\section{Datasets}\label{app:dataset}

We briefly introduce the particulars of each dataset as follows: 
% \begin{enumerate}
\noindent \textbf{NExT-QA} \cite{xiao2021next} is a manually annotated dataset that benchmarks the causal and temporal object interaction reasoning. Its videos are sourced from VidOR \cite{shang2019annotating,shang2019relation} and cover various daily activities. By default, each question has 5 options with 1 correct answer and 4 distractor answers.

\noindent \textbf{TGIF-QA} \cite{jang2017tgif} challenges repeating action recognition and temporal state transition. The videos are short GIFs and are trimmed to contain only the content of interested for the paired questions. It provides 5 options for each question and requires the models to pick the correct one. \textbf{TGIF-QA-R} \cite{peng2021progressive} derives from TGIF-QA action and state transition tasks by fixing the answer bias issue. Therefore, It shares the same videos, questions and correct answers with TGIF-QA. In our experiment, we further remove the redundant candidate answers in TGIF-QA-R for better evaluation.

\noindent \textbf{STAR-QA} \cite{wu2021star} benchmarks situated visual reasoning. It is based on Action Genome \cite{ji2020action} to curate questions and answers that verify a wide range of reasoning capabilities about human-object interaction, temporal sequence analysis, action prediction, and feasibility inference. Its videos feature single person indoor activities. 

\noindent \textbf{Causal-VidQA} \cite{li2022representation} goes beyond visual evidence reasoning to study visual commonsense in videos. It sets four type of questions: Description, Explanation, Prediction and Counterfactual ones. The tasks are defined as multi-choice selection. In particular, a correct prediction for the questions in prediction and counterfactual require \emph{both} the answers and the corresponding reasons to match the ground-truth ones. 

\noindent \textbf{TGIF-FrameQA} is a sub-task defined in the TGIF-QA dataset \cite{jang2017tgif}. It mimics ImageQA \cite{antol2015vqa} by posing questions that invoke a single frame for answer. The QA pairs in \textbf{MSRVTT-QA} \cite{xu2017video} are automatically generated from video descriptions \cite{xu2016msr}. The two datasets focus on the recognition of the video objects/attributes/actions/activities; their questions rarely invoke temporal relations. 

\section{Implementation Details}
\label{app:imp}
For experiments on STAR \cite{wu2021star}, we obtain the video segments associated with each QA pair according to the given time stamps; the segments are then treated as independent videos for processing. Particularly, we only use the QA annotations for training following the standard in video question answering. In addition, the final submission to the evaluation server is obtained by training with both the train and validation data following the original paper \cite{wu2021star}. For Causal-VidQA, we process the videos in the same way as other datasets and ground-truth visual object annotations are not used. For TGIF-Action and Transition tasks, we find that adding randomness to the multiple choices in self-supervised contrastive learning will hurt the performance, and thus we do not use it.

For contrastive learning on target datasets, we obtain the hard negative answers and questions according to the question types. While the question types maybe provided in some datasets, we simply obtain them according to the starting three question words\footnote{why, what, where, which, who, how (many)/(times) $+$ is(are)/does(do).} for adaptability. Moreover, we use the same rule for all datasets. We find that this works well though the obtained question types may not be perfect. For specific training, we first train the whole model end to end, and then freeze the language model to fine-tune the other parts of the best model obtained at the $1st$ stage. The best results obtained in the two stages are determined as final results. 
For pretraining with weakly-paired video-text data \cite{bain2021frozen}, we preprocess the videos in the same way as for QA (\ie~ samples 8 clips and 10 regions per frame) and pretrain the model with an initial learning rate of $5\times10^{-5}$ and batch size 64. Besides, a text token is corrupted at a probability of 15\% in masked language modelling. Following \cite{liu2019roberta,yang2021just}, a corrupted token will be replaced with 1) the `[MASK]' token by a chance of 80\%, 2) a random token by a chance of 10\%, and 3) the same token by a chance of 10\%. We train the model by maximal 2 epochs which gives to the best generalization results. Our pretraining costs about 2 hours on 4 Tesla V100 GPUs.

\section{Result Analysis}
\subsection{Results Per Question Type on NExT-QA}
\label{app:comp}
We compare CoVGT with the recent methods that have reported accuracy per question type on NExT-QA. As shown in Tab.~\ref{tab:resty}, our methods outperforms the competitors significantly in answering all type of questions except for those in the counting group. The weaker performance on counting could be majorly due to our sparse sampling, \eg~32 frames per video and 5 regions per frame. In addition, the smoothing effect of attention (either Transformer or GNN) could also jeopardize the counting performance.
% Notably, our improvements are stable across questions that are in the majority and  minority.

\setlength{\tabcolsep}{8pt}
\begin{table*}[t]
\begin{center}
\small
\caption{Results per question type on NExT-QA val set.}
\label{tab:resty}
\vspace{-0.5em}
\scalebox{0.9}{
\begin{tabular}{lccccccccccc}
\toprule
\multirow{2}{*}{Methods} & \multicolumn{3}{c}{Acc@C} & \multicolumn{3}{c}{Acc@T} & \multicolumn{4}{c}{Acc@D} & \multirow{2}{*}{Acc@All}\cr
\cmidrule(lr){2-4} \cmidrule(lr){5-7} \cmidrule(lr){8-11}
 & \tabincell{c}{Why\\(36\%)} & \tabincell{c}{How\\(12\%)} & \tabincell{c}{Overall\\(48\%)} & \tabincell{c}{Prev\&Next\\(16\%)} & \tabincell{c}{Present\\(13\%)} & \tabincell{c}{Overall\\(29\%)} & \tabincell{c}{Count\\(4\%)} & \tabincell{c}{Location\\(5\%)} & \tabincell{c}{Other\\(7\%)} & \tabincell{c}{Overall\\(16\%)} \cr

\midrule
HGA \cite{jiang2020reasoning} & 46.99 & 44.22 & 46.26 & 49.53 & 52.49 & 50.74 & 44.07 & 72.54 & 55.41 & 59.33 & 49.74 \cr
TrajG \cite{xiao2022rethinking} & 52.81 & 47.44 & 51.40 & 51.11 & 53.70 & 52.17 & 46.89 &75.25 & 58.03 & 62.03 & 53.30 \cr
P3D-G \cite{cherian2022} & 52.39 & 48.36 & 51.33 & 50.91 & 54.28 & 52.30 & 46.02 & 77.08 & 58.31 & 62.58 & 53.40 \cr
\midrule 
CoVGT & \underline{59.77} & \underline{56.08} & \underline{58.80} & {\bf56.16} & \underline{59.28} & \underline{57.44} & {\bf50.85} & 
{\bf82.71} & \underline{67.21} & \underline{69.37} & \underline{60.01} \cr
CoVGT(PT) & {\bf60.65} & {\bf56.95} & {\bf59.69} & \underline{56.06} & {\bf60.78} & {\bf58.00} & \underline{47.46} & \underline{82.37} & {\bf70.82} & {\bf69.88} & {\bf60.73} \cr 
\bottomrule
\end{tabular}
}
\vspace{-0.4cm}
\end{center}
\end{table*}

\subsection{Contrastive Learning}
\label{app:cst}
\begin{figure*}[t!]
  \begin{center}
    \includegraphics[width=1.0\textwidth]{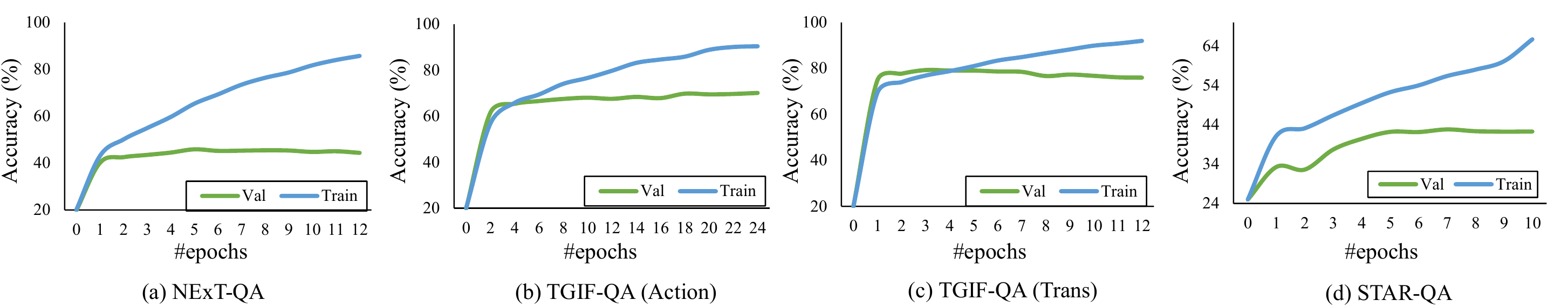}
  \end{center}
   \vspace{-0.5em}
  \caption{Accuracy with regard to the training epochs.}
  \vspace{-0.4cm}
  \label{fig:of}
\end{figure*}
\textbf{Contrastive Learning \vs Classification}. 
To further study whether the poor performance of the classification model variant (Sec.~\ref{sec:cst}) is caused by the cross-modal transformer or the classification layer. We keep our cross-modal interaction mechanism (introduced in Sec.~\ref{sec:cm}) and discard the cross-modal transformer. Thus, we directly send the cross-modal interacted video representation $f^{qv}$ to a classification layer for multi-choice classification.
There are $|\mathcal{A}_{mc}|$ such video representations obtained by interacting the video with different question-answer pairs, and each is mapped to a scalar that gives the probability of the candidate answer to be a correct one. The results in Tab.~\ref{tab:cls} show that this implementation variant (CM) wins slightly on the cross-modal transformer (CMTrans). The small improvements could be attributed to that our cross-modal interaction do not have self-attention weights (single modality) which usually take large portion of the attention distribution in cross-modal transformers. Thus, our cross-modal interaction focuses more on cross-modal information exchange, and thus benefits the final results. However, the small improvement does not change the fact that a classification setting is sub-optimal for VideoQA, at least in the pretraining-free setting together with advanced language models.
\setlength{\tabcolsep}{4pt}
\begin{table}[t!]
    \small
    \centering
    \caption{A detailed study of the classification model variants.
    }
    \label{tab:cls}
    \vspace{-0.5em}
    \scalebox{0.9}{
    \begin{tabular}{l|ccc|c}
    % \hline\hline
    \multirow{2}*{Models} &  \multicolumn{4}{c}{NExT-QA Val} \\ \cline{2-5}
    ~ & Acc@C & Acc@T & Acc@D & Acc@All \\ 
    \hline
    CMTrans ([CLS]) & 42.96 & 46.96 & 53.02 &  45.82 \\ 
    CM (CLS) &  {\bf44.46} & {\bf47.33} & {\bf54.70} & {\bf46.98} \\
    % \hline
    \end{tabular}
    }
    % \vspace{-0.4cm}
\end{table}

\subsection{Study of QA Short-Cut in Open-ended QA}
\label{app:shortcut}
Tab.~\ref{tab:oeqa} shows that the QA short-cut in Eqn.~\eqref{equ:oe} does contribute to the performance. The results suggest that we can take advantage of the language biases for better performance on the test data. 
\setlength{\tabcolsep}{3pt}
\begin{table}[t!]
    \small
    \centering
    \caption{Study of the QA short-cut prediction.}
    \label{tab:oeqa}
    \vspace{-0.5em}
    \scalebox{1.0}{
    \begin{tabular}{l|c|c}
    % \hline\hline
  Methods & TGIF-FrameQA & MSRVTT-QA \\
    \hline
    VGT & {\bf61.6} & {\bf39.7} \\ 
    VGT w/o QA & 61.2 & 39.1 \\ 
    % \hline
    \end{tabular}
    }
    % \vspace{-0.4cm}
\end{table}

\subsection{Study of Finetuning}
\label{app:adapt}
\textbf{QA Adaptation}. We additionally study fine-tuning VGT by converting the QA-pairs into descriptions so as to reduce the data difference between pretrain and finetune. In our implementation, we convert the question whose pattern is clear and easy to adapt, otherwise we directly concatenate the answer behind the question with a special token `[SEP]'. For instance,\texttt{"why is the baby crying? fell backwards"} to \texttt{"the baby crying [SEP] fell backwards"}, and \texttt{"what did the man do after squatting down? wash hands."} to \texttt{"the man [SEP] wash hands [SEP] after squatting down"}. Our results in Tab.~\ref{tab:pt} (Adapt All) show that such adaptation benefits the performances on causal questions but hurts the performance of others. We speculate the main reason is that the converted descriptions of a given sample are quite similar for temporal and descriptive questions since the candidate answers are short but they share the same long question. As a result, the descriptions render it hard to disambiguate between the correct and incorrect answers. Based on such observation, a better alternative is to only adapt the questions in the causal group. We can see that the overall performance improves on the validation set (Adapt C). Nevertheless, we find that model does not generalize well to the test set, \ie, 55.08\% which is worse than 55.7\% obtained by the model without considering adaption. As the experiment is not our focus, we stop here and hope that our pioneer attempt can spark more effective and interesting works.

\setlength{\tabcolsep}{6pt}
\begin{table}[t!]
    \small
    \centering
    \caption{Study of fine-tuning.
    }
    \label{tab:adapt}
    \vspace{-0.5em}
    \scalebox{0.9}{
    \begin{tabular}{l|ccc|c}
    % \hline\hline
    \multirow{2}*{Methods} &  \multicolumn{4}{c}{NExT-QA Val} \\ \cline{2-5}
    ~ & Acc@C & Acc@T & Acc@D & Acc@All \\ 
    \hline
    FT (baseline) &  53.43 & {\bf56.39} & 69.50 & 56.89 \\ 
    \hline
    FT Adapt All &  {\bf54.89} & 54.40 & 67.57 & 56.69 \\
    FT Adapt C &  {\bf54.89} & 55.15 & {\bf69.76} & {\bf57.29} \\ 
    % \hline
    \end{tabular}
    }
    % \vspace{-0.4cm}
\end{table}

\begin{figure}[t!]
  \begin{center}
    \includegraphics[width=0.45\textwidth]{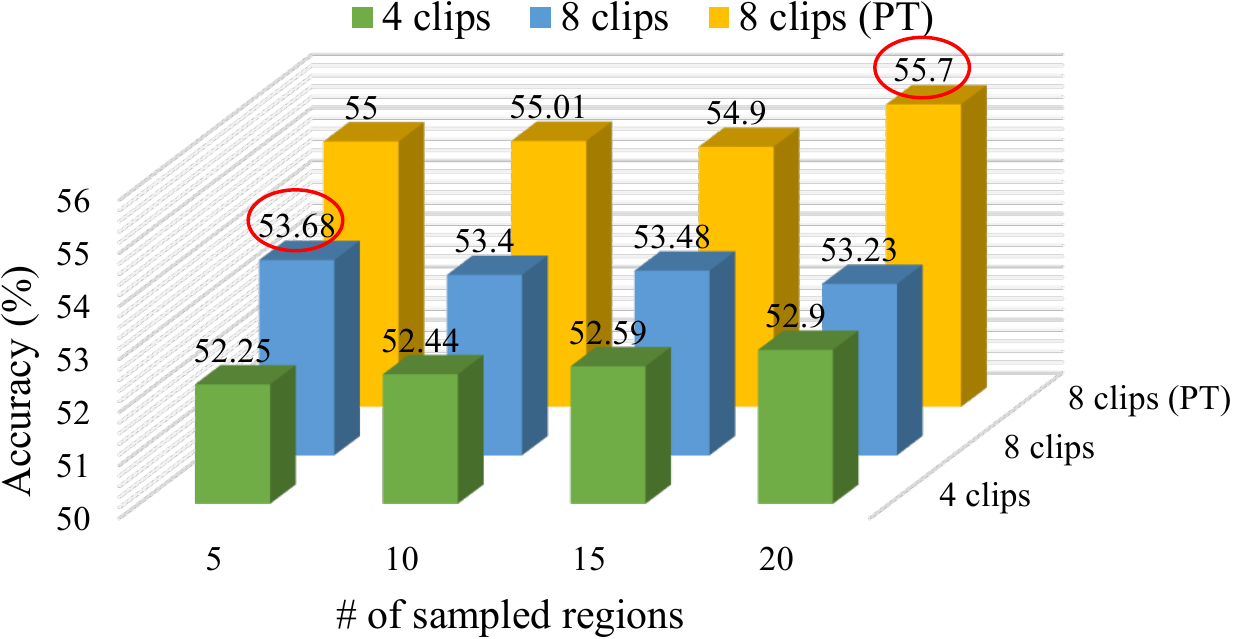}
  \end{center}
  \vspace{-0.5em}
  \caption{Investigation of sampled video clips and region proposals per frame. Results are reported on NExT-QA test set.}
  \label{fig:region}
  \vspace{-0.4cm}
\end{figure}

\begin{figure*}[t!]
\centering
    \includegraphics[width=1.0\textwidth]{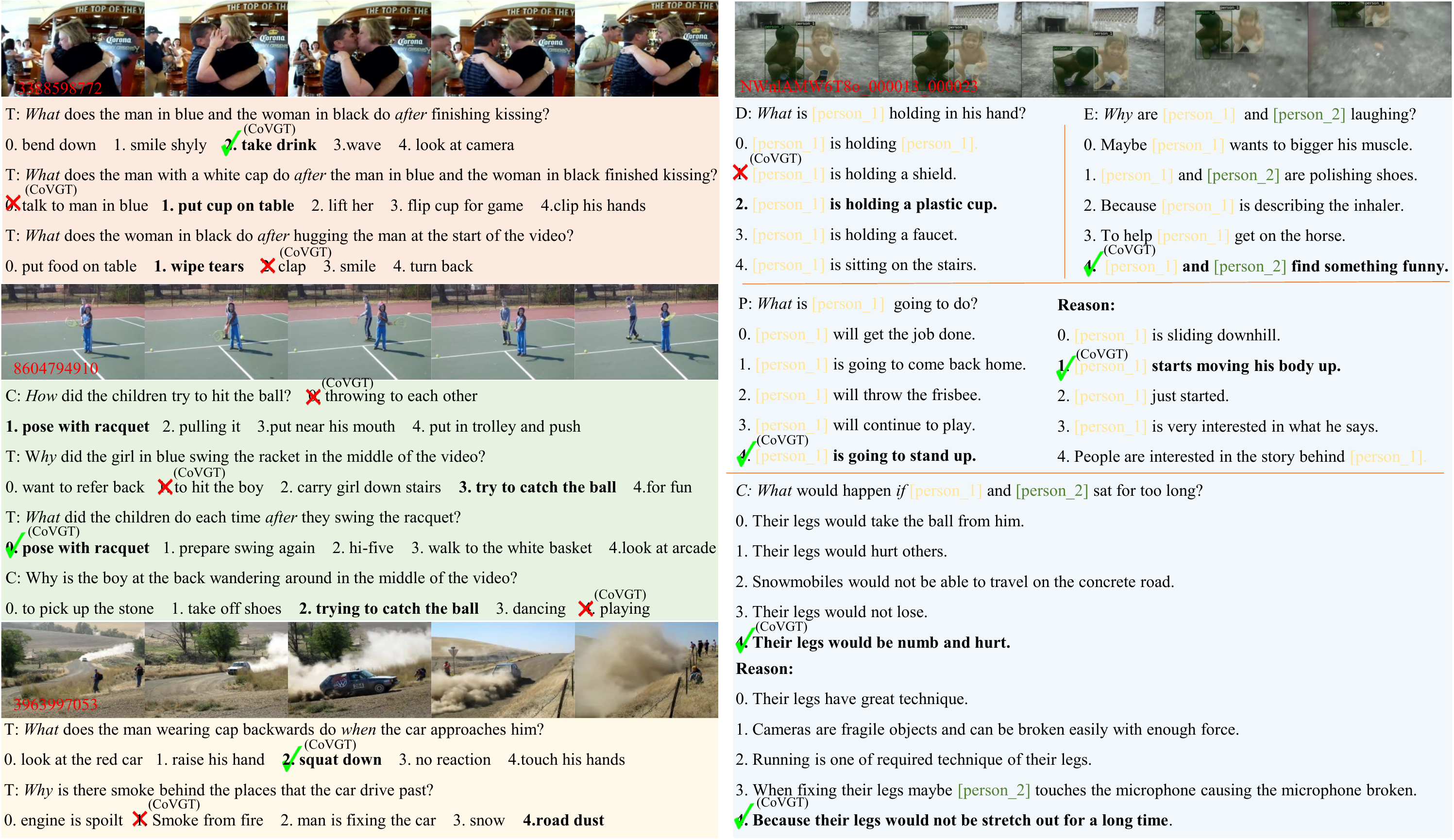}
  \caption{Prediction results on (left) NExT-QA \cite{xiao2021next} and (right) Causal-VidQA \cite{li2022representation}. Ground-truth answers are highlighted in \textbf{bold}. Note: The masks and bounding boxes are attached in the video from Causal-VidQA and not our detection.}
  \label{fig:appvis}
\vspace{-0.4cm}
\end{figure*}
\subsection{Video Sampling}
\label{app:vsp}
In Fig.~\ref{fig:region}, we study the effect of sampled video clips and region proposals on NExT-QA \cite{xiao2021next} based on the VGT model. Regarding the number of sampled video clips, we find that the setting of 8 clips steadily wins on 4 clips. This is understandable as the videos in NExT-QA are relatively long. As for the sampled regions, when learning the model from scratch, the setting of 5 regions gives relatively better result, \eg, 53.68\%. Nonetheless, when pretraining are considered, the setting of 20 regions gives better result, \eg, 55.70\%. Such difference could be due to that learning with more regions can yield over-fitting issues when the dataset is not large enough, since the constructed graph become much larger and more complex. Our speculation is also supported by the observation that the accuracy increases with the number of regions when we only sample 4 video clips (less graph nodes). Based on the observations, we use 5 regions in the pretraining-free experiments and 10 regions in the \emph{pretrain and finetune} experiments for CoVGT to balance between accuracy and efficiency. 
\subsection{More Qualitative Analysis}
\label{app:qua_res}
We show some of our prediction results in Fig.~\ref{fig:appvis}. The examples from NExT-QA suggest that understanding the fine-grained video object interactions is of great challenge. Although we have remarkably improved the previous SOTA results, we find that our model still fails to answer a large number of questions. The failure cases show that the model can answer most of the questions reasonably, but the answers are irrelevant to the factual video contents. The observation indicates that our model can well understand the questions yet still not strong enough in comprehending the fine-grained video contents or finding the cross-model correspondences between vision and text. On the other hand, the example from Causal-VidQA shows that our model can successfully answer the questions that emphasize temporal dynamics and find the reasonable explanations as well, such as ``going to stand up'' and ``start moving his body up''. 

\end{document}